%% file: main.tex
\documentclass[pdflatex,sn-nature,a4paper]{sn-jnl}  

\usepackage{a4wide}
\usepackage{graphicx}%
\usepackage{amsmath,amssymb,amsfonts}%
\usepackage[title]{appendix}%
\usepackage{todonotes} 
\usepackage{bbm} 
\usepackage{numprint} 
\usepackage{makecell}

\usepackage[utf8]{inputenc}   
\usepackage[T1]{fontenc}      

\begin{document}

\renewcommand{\appendixname}{Supplementary Information}

\title[Article Title]{SpecTUS: Spectral Translator for Unknown Structures annotation from EI-MS spectra}

\author[1]{\fnm{Adam} \sur{Hájek}}\email{ahajek@mail.muni.cz}
\author[2]{\fnm{Michal} \sur{Starý}}\email{michal.stary@tum.de}
\author[3]{\fnm{Elliott} \sur{Price}}\email{elliott.price@recetox.muni.cz}
\author[4,5]{\fnm{Filip} \sur{Jozefov}}\email{filip.jozefov@uochb.cas.cz}
\author[3]{\fnm{Helge} \sur{Hecht}}\email{helge.hecht@recetox.muni.cz}

\author*[1]{\fnm{Aleš} \sur{Křenek}}\email{ljocha@ics.muni.cz}

\affil[1]{\orgdiv{Institute of Computer Science}, \orgname{Masaryk University}, \orgaddress{Šumavská 525/33}, \\ \city{Brno}, \postcode{602~00}, \country{Czech Republic}}
\affil[2]{\orgdiv{School of Computation, Information and Technology}, \orgname{Technical University Munich}, \orgaddress{Boltzmannstraße 3}, \city{Garching bei München}, \postcode{85748}, \country{Germany}}
\affil[3]{\orgdiv{RECETOX}, \orgname{Masaryk University}, \orgaddress{Kamenice 753/5}, \city{Brno}, \postcode{625~00}, \country{Czech Republic}}
\affil[4]{\orgdiv{Faculty of Informatics}, \orgname{Masaryk University}, \orgaddress{Botanická 554/68a},\\  \city{Brno}, \postcode{602~00}, \country{Czech Republic}}
\affil[5]{\orgname{Institute of Organic Chemistry and Biochemistry of the CAS}, \orgaddress{Flemingovo nám. 542/2}, \city{Praha 6-Dejvice}, \postcode{160~00}, \country{Czech Republic}}

\abstract{
Compound identification and structure annotation from mass spectra are essential in drug detection, forensics, and small molecule discovery. All the current approaches to compound identification from electron ionization mass spectra (EI-MS) are dependent on different forms of searching databases that are orders of magnitude smaller than the space of potential molecular structures they attempt to cover. 

We introduce SpecTUS: Spectral Translator for Unknown Structures, a deep learning model for \textit{de novo} structural annotation, translating gas chromatography EI-MS spectra directly into molecular structures without requiring reference databases. This enables the identification of novel compounds absent from spectral libraries.

In a rigorous evaluation, SpecTUS significantly outperformed standard database search techniques. On a held-out test set of \numprint{28267} spectra from NIST 20, the model's single suggestion perfectly reconstructed 43\% of the subset's compounds, strictly outperforming hybrid database search (common method among practitioners) in 76\% of cases. With ten suggestions, it achieved 65\% perfect reconstructions, surpassing hybrid search in 84\% of cases.}


\maketitle


Gas chromatography-mass spectrometry (GC-MS) is a widely used method for identifying compounds, particularly those that are volatile and thermally stable. In the first stage, compounds are separated via gas chromatography. Next, neutral molecules are ionized, generating charged ions for analysis. The MS then measures the mass-to-charge ratio (m/z) of each fragment, along with the abundance of fragments at each m/z value. After processing, this analysis produces a mass spectrum for each compound, represented as a series of peaks. Each peak corresponds to a unique fragment mass and is encoded as a pair of m/z value and its relative intensity, reflecting the fragment's abundance in the sample. 

Different mass spectrometry methods produce spectra with distinct characteristics based on, amongst other factors, the ionization technique and the number of spectrometry iterations. One of the most widely used, established methods is GC-EI-MS, where EI stands for electron ionization. A standard electron energy of 70 eV has been used for decades, ensuring relative consistency across spectra and compatibility of different spectral libraries. However, the variation in spectra still persists \cite{KELLY2018_gcmc_variability}.  As this approach is central to our research, GC-EI-MS will be the focus of this study.

Compound identification from GC-EI-MS spectra has been approached in three main ways: \textit{database search, extended database search,} and \textit{de novo generation.}

In a standard database search, we compare the spectrum against a library of experimentally measured reference spectra using either simple similarity search (SSS) or hybrid similarity search (HSS)~\cite{hss_gcms_2017,hybrid_search}. SSS, which matches only peaks with the same (similar) m/z value, is effective when the compound being analyzed is present in the spectral library, a scenario known as the \textit{spectral match} task. When the compound is absent from the library -- the \textit{closest structure identification} task -- HSS typically performs better. HSS incorporates structural information by matching neutral losses in addition to fragment ions. It calculates these neutral losses by shifting one of the spectra based on the molecular weight difference (DeltaMass) between the query and library compounds. However, for HSS to work accurately, a precise DeltaMass is required, which depends on the molecular weight of the query compound — a value that is not always reliably estimated from GC-EI-MS spectra.
A~widely used GCMS-ID webserver~\cite{gcms-id} implements similar search approach combined with scoring on retention index;
together with an impressive collection of curated references spectra 
it represents the state-of-the-art in standard database search.

Extended search approaches aim to address the limited size of spectral databases by expanding the scope of references available for comparison.
One extreme from among these methods is an \emph{ab-initio} generation of synthetic EI-MS spectra, as implemented in QCEIMS~\cite{qceims},
by simulating the fragmentation with quantum-chemical calculation.
However, achieving sufficient accuracy requires high-level methods (e.g., DFT), making this approach impractical for large-scale database generation~\cite{eimspred}.


Another approach to EI spectra generation is rule-based prediction~\cite{Schymanski2009ruleBasedComparison}. In this family of approaches commercial software such as MS-Fragmenter~\cite{ACD_MS_Fragmenter} exists, but those lack open source code and algorithm transparency.

Machine learning-based methods offer a faster alternative for generating synthetic spectra, though often at the cost of losing some accuracy. NEIMS~\cite{neims} algorithmically computes the extended connectivity fingerprints (ECFP)~\cite{ecfp} of a molecular structure and feeds it into a~smartly designed multilayer perceptron to generate a low-resolution synthetic EI-MS spectrum. The model, trained on \numprint{240000} experimental spectra from the NIST 17 dataset, is known for its simplicity and reasonable reliability, making it the foundation of several other methods~\cite{deepei, fastei}. A more sophisticated alternative, RASSP~\cite{rassp}, 
processes the structural formula augmented by atomic features with a~graph neural network (GNN).
The GNN output is passed to an attention-equipped neural network to predict fragments and their relative abundances. These fragments are then transformed into an EI-MS  spectrum using exact atomic masses and isotopic patterns, producing high-resolution outputs. Compared to NEIMS, RASSP generates more accurate spectra but is limited to a small set of atom types (8 in the published version) and by the number of possible fragment formulae (4096).

To improve accuracy and search speed in extended spectral databases, recent methods leverage machine learning to map EI-MS spectra into alternative vector spaces. DeepEI~\cite{deepei} is a model trained to translate spectra into molecular fingerprints
which can be rapidly computed for vast structural libraries such as ZINC or PubChem, moving the similarity search into the space of fingerprints. FastEI~\cite{fastei} combines NEIMS with a spectral embedding method inspired by Word2Vec~\cite{word2vec}, producing vector representations (embeddings) of spectra that later serve for rapid library searching. 

To address the challenges of coverage and information loss from intermediate translations, recent \textit{de novo} methods directly translate spectra into molecular structures. Inspired by neural machine translation architectures, these methods typically generate SMILES~\cite{Weininger1988SMILES} strings autoregressively based on encoded spectral information. 

So far, all the \textit{de novo} models have been exclusively trained on tandem mass spectrometry spectra (LC-MS/MS) resulting from a two-staged fragmentation process. In the first stage, the method uses soft ionization process (ESI) that enables extracting the precursor ion mass (approximately the molecular mass of the analyzed compound). The precursor ion is further fragmented by hard ionization process in the second stage yielding a mass spectrum similar, yet not comparable, to EI-MS.

For instance, MassGenie~\cite{massgenie}, a $\sim$400 million parameter encoder-decoder transformer, maps MS/MS spectra, represented as embeddings of m/z values combined with positional encoding into SMILES strings. It was pretrained on 4.7 million synthetic spectra and finetuned on approximately \numprint{200000} experimental spectra from the GNPS dataset. Another method, Spec2Mol~\cite{spec2mol}, separately trains a GRU autoencoder on 138 million SMILES strings, creating a latent space of molecular structures. Further, the trained GRU decoder is combined with a CNN encoder that learns to map experimental MS/MS spectra to the GRU's molecular latent space using \numprint{28000} examples from the NIST library. MSNovelist~\cite{msnovelist} employs an RNN encoder-decoder architecture relying on SIRIUS~\cite{sirius} to predict a~molecular formula and structural fingerprint from MS/MS spectra. The encoder maps these inputs onto a~latent space vector used further by an LSTM decoder to generate a~SMILES string character-by-character, aided by a continuously updated molecular subformula representing remaining atoms. The model was trained on a~dataset combining HMDB, COCONUT, and DSSTox databases comprising 1.2 million molecular structures. Another method, MS2Mol~\cite{ms2mol}, employs an encoder-decoder transformer enhanced with precursor mass information trained on around 1 million MS/MS spectra. The model emphasizes output ordering optimization with a dedicated re-ranking model. Finally, Mass2SMILES~\cite{mass2smiles} utilizes a smaller transformer encoder paired with a TCN decoder and explores variational autoencoders with continuous latent space generation trained on \numprint{83000} experimental MS/MS spectra from the GNPS dataset. 

While tandem MS enables richer analysis and can yield precursor ion weights or even formulas of compounds, the method’s increased complexity introduces more adjustable parameters. Due to the lack of consensus on standard settings, its variability can reduce consistency across spectral libraries.


We argue that despite missing precursor ion weight, the character and consistency of EI-MS spectra make them a well-suited representation for machine learning models focused on molecular structure annotation. In this paper, we introduce SpecTUS: a \emph{Spectral Translator for Unknown Structure reconstruction from low-resolution EI-MS spectra}. At its core, SpecTUS is a 354 million parameter encoder-decoder transformer model that takes encoded mass spectrum as input and directly generates molecular structure as a~SMILES string. The model can produce an arbitrary number of candidate structures, each accompanied by an inherent certainty score.

Our training process began with pretraining on a synthetic dataset of $2\times8.6$ million spectra generated by NEIMS and RASSP models, believed to capture sufficient foundation knowledge on the structure-spectra relationships.
We then finetuned SpecTUS on \numprint{232025} clean, experimentally measured spectra from NIST 20 (NIST/EPA/NIH Mass Spectral Library of electron ionization spectra, 2020 version) to adjust to true experimental data. Additionally, we conducted experiments to assess the impact of various pretraining strategies, tokenization schemes, and input encoding methods, adapting the model to suit the spectra-to-molecules translation task as closely as possible. Our evaluation demonstrates that SpecTUS effectively identifies unseen compounds, showcasing its potential for accurate, rapid structure annotation from EI-MS spectra. 

Alongside the paper, we release two datasets together containing 17.2 million synthetic spectra generated by NEIMS and RASSP models, the pretrained SpecTUS model, and all necessary training and evaluation scripts, accompanied by a detailed tutorial. While the final finetuned model cannot be distributed freely due to NIST's commercial licensing, we provide preprocessing scripts and our data splits to support full reproducibility. We also provide a hosted demo application with limited access to the final model for anyone to test the model's capabilities.

\section*{Results}
\label{s:results}

SpecTUS evaluation was driven by the motivation to demonstrate that the method is capable of \textit{generalization},
i.e. it captures knowledge contained in the training data and is able to apply it to unseen spectra of unseen compounds.

Moreover, we show that SpecTUS outperformed traditional database search methods when identifying compounds not present in the reference database. This highlights its capability to address the limitations of existing approaches in the compound identification task.

\subsection*{Baselines}
\label{s:baselines}
We compared SpecTUS against database search methods using the NIST 20 dataset, where the SpecTUS training set (\numprint{232035} spectra, see Sect.~\nameref{s:finetuning-dataset} for details) served as the reference library for database search and disjoint testing sets were used as queries. The reference library did not overlap with any test or validation data, and it remained representative for practical applications in terms of size.

Our evaluation included the following baselines:
\begin{itemize}
    \item \textbf{Simple Similarity Search (SSS)}:
This method is universally applicable for database searches as it does not rely on any additional measurements or precise molecular weight (MW) estimates. It matches query spectra to the library based purely on spectral similarity, computed as augmented cosine similarity on a~set of matching peak pairs~\cite{hybrid_search}.
    \item \textbf{Hybrid Similarity Search (HSS)}:
The most commonly used method for closest structure identification from a database. HSS combines fragment ion matching with neutral loss analysis and incorporates the molecular weight of the query compound into the computation~\cite{hybrid_search}. While effective, HSS depends on knowing the molecular weight information, which is not always reliably determined from EI-MS spectra.
However, for our evaluation purposes, we provide the weight available in the testing sets, which biases the results towards better HSS performance compared to the real-world scenario.
    \item \textbf{Best Database Candidate (BDC)}:
For this baseline, the best candidate is selected from the reference library based on structural similarity between the query compound and reference compounds, measured by Tanimoto similarity over Morgan fingerprints~\cite{ecfp}.
This approach is artificial, and we don't have access to it in real-world scenarios because the analyte's structure is precisely what we aim to determine.
We use BDC to determine the upper bound for database search methods.
\end{itemize}
These baselines reflect both standard practices and theoretical limits of conventional methods for spectra identification.

\subsection*{Metrics}
Evaluation of SpecTUS performance, its comparison with the baseline, as well as with other solutions is based on
several metrics we describe briefly in this section.  
Detailed formal definitions of those can be found in Sect.~\nameref{s:methods}.
All the metrics assume they are applied on a~given test set of spectra for which the structures (ground truth) are known.

For a~given set of test spectra, 
\textbf{top-$k$ accuracy} ($\text{Acc}_k$) lets the model predict $k$ 
structures for each spectrum, and measures the proportion of spectra for which the correct structure appears among the model's $k$ predictions.

\textbf{Average top-$k$ similarity} (Sim$_k$) quantifies how structurally similar, on average, the most accurate candidate is to the ground truth molecule. For each query, the similarity is calculated as the highest Tanimoto similarity between the Morgan fingerprints of the ground truth molecule and each of the first $k$ retrieved candidates. The similarities are then averaged over all queries in the testing set. 

To compare two methods, we define the \textbf{win rate} of the first method as the proportion of test spectra for which its best prediction among the top $k$ is strictly better than that of the other method. To overcome the imperfection of Morgan fingerprints, an exact structural match is considered the best outcome; if neither method achieves an exact match, the one with the higher Tanimoto similarity on Morgan fingerprints is deemed superior.
Similarly, the \textbf{at-least-as-good rate} complements the win rate. That is, draws are counted in favor of the first method too.

\subsection*{Scenarios}
We evaluated the model's performance in three scenarios, varying the number of returned candidates: 1, 10, and 50. As the number of candidates increased, the overall top-$k$ performance improved, demonstrating the model's ability to provide more comprehensive predictions. However, while the model assigns sequence probabilities to rank its generated candidates, this ranking is not sufficiently reliable to replace expert evaluation. For instance, the model's top-ranked candidate from a set of 10 is, on average, slightly less accurate than the one candidate generated in the single-candidate scenario. This suggests that manual expert validation remains crucial for multi-candidate scenarios. Consequently, we did not use the model-based ranking during evaluation. The three evaluated scenarios are as follows:
\begin{itemize}
    \item \textbf{Single-candidate scenario (SpecTUS$_1$)}: Requires no manual intervention but has lower accuracy, making it suitable for high-throughput workflows where speed is prioritized over precision.
    \item \textbf{Ten-candidate scenario (SpecTUS$_{10}$)}: Offers a balanced trade-off, providing sufficient diversity in predictions while keeping manual evaluation manageable. This makes it particularly well-suited for practical use in analysis workflows.
    \item \textbf{Fifty-candidate scenario (SpecTUS$_{50}$)}: Demonstrates the model's potential for providing highly accurate predictions but at the cost of substantial manual effort to evaluate all candidates.
\end{itemize}
By tailoring the number of generated candidates to the specific application, users can leverage the model's strengths while balancing accuracy and effort.

\subsection*{Datasets}
To evaluate SpecTUS, we used a held-out NIST 20 test split (see Section \ref{s:finetuning-dataset}) along with several publicly available libraries, including SWGDRUG, Cayman, and MONA. The NIST 20 and SWGDRUG databases provide highly curated spectra, making them ideal for training and validation. In contrast, testing on the Cayman and MONA libraries allowed us to assess the model’s performance on spectra collected and processed with less standardized methods, highlighting the challenges posed by varying spectral acquisition techniques.

To demonstrate the generalization capabilities of SpecTUS, we carefully filtered all test sets to exclude any compounds present in the synthetic or experimental training datasets. Additionally, compounds containing deuterium, which was not included in the training datasets, were removed, and model-specific filters were applied -- a maximum m/z value of 500, a maximum of 300 peaks per spectrum, and a maximum SMILES length of 100 characters. The vast majority of spectra removed from SWGDRUG, Cayman library, and MONA were excluded due to overlap with the NIST 20 training set.

Table~\ref{t:test_set_size} provides a detailed overview of the final dataset sizes before and after filtering.

\begin{table}
\caption{The number of spectra in each test library before and after filtering. The filtering included removing overlaps with the NIST training set, removing compounds containing deuterium, and filtering out datapoints with m/z exceeding 500, more than 300 peaks, or SMILES string exceeding 100 characters.}
\label{t:test_set_size}
\begin{tabular}{r|r|r|r|r}
  & NIST test & SWGDRUG & Cayman & MONA \\
\hline
original size & 29218 & 3589 & 2262 & 18914 \\
used for testing & 28267 & 1640 & 469 & 5015 \\
\end{tabular}
\end{table}

\subsection*{Baseline results}

\begin{figure}
    \centering
    \includegraphics[width=0.45\textwidth]{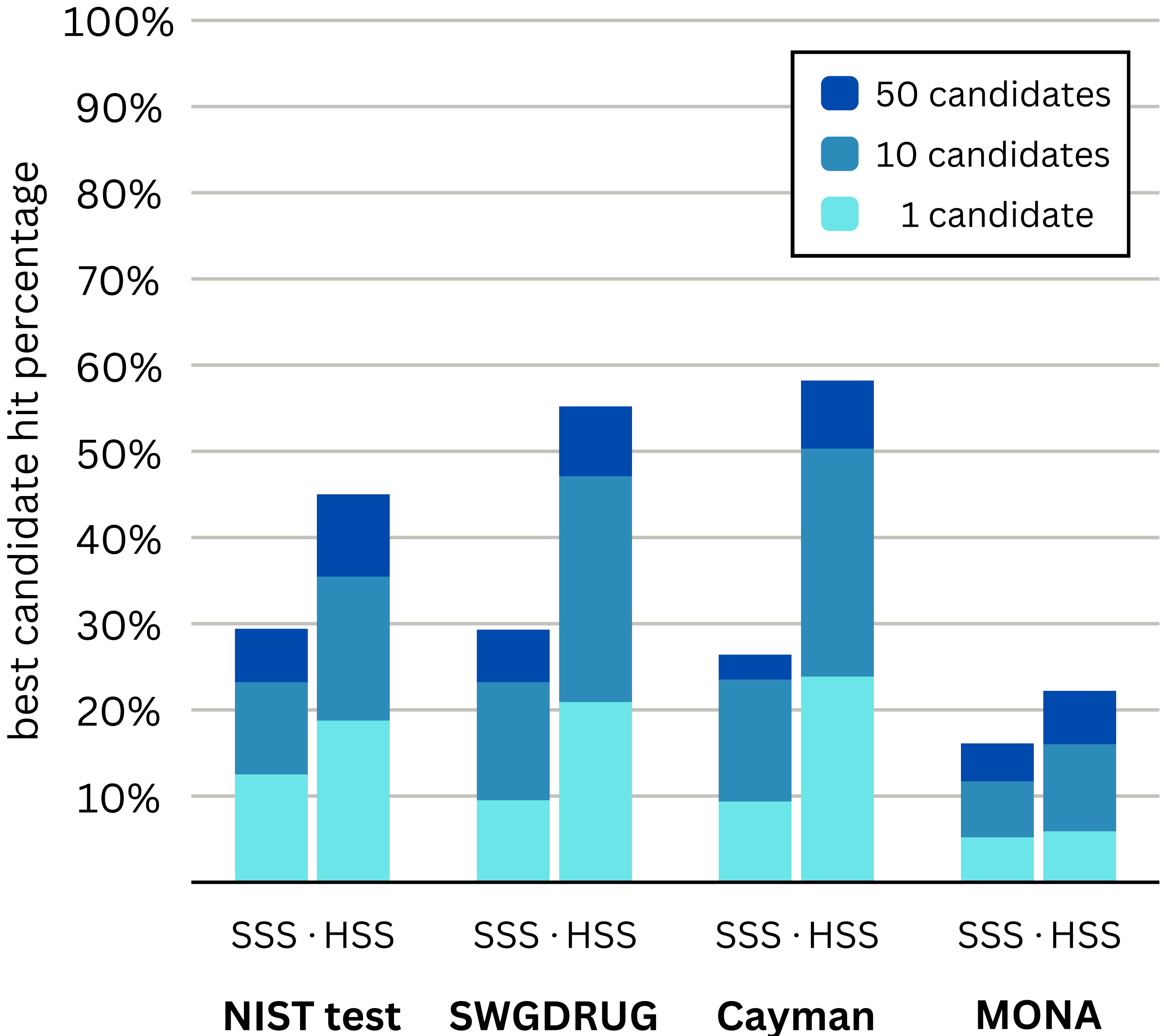}
    \caption{Percentage of cases where database search methods (SSS and HSS) successfully retrieved the closest structure from the reference database among the top-1, top-10, and top-50 suggested candidates. Performance is evaluated across all test sets: NIST test split, SWGDRUG, Cayman, and MONA.}
    \label{fig:db_comparison} 
\end{figure}

Initially, we compared SSS and HSS, to the upper-bound based on structural similarity (BDC) to quantify the limitations of current database search methods (Figure~\ref{fig:db_comparison} and Table~\ref{tab:db_search_comparison}). The analysis focuses on the percentage of cases where the standard database search methods (SSS, HSS) successfully retrieved the most structurally similar compound from the database among their top-$k$ candidates.  Importantly, this metric does not evaluate the quality of the retrieved candidates relative to the ground truth; rather, it highlights how standard database search methods perform within the inherent constraints of database searching. Across all test sets (NIST test, SWGDRUG, Cayman, and MONA), several patterns emerged.

The performance of database search methods was relatively consistent across datasets, except for MONA, 
possibly due to higher occurrence of TOF spectra or generally less thorough curation. Surprisingly, results for the NIST test set were lower than those of Cayman, despite  better  curation of the NIST library.

HSS consistently outperformed SSS, demonstrating its superiority across all datasets. However, even HSS struggled as the top-50 candidate list did not surpass the 60\% success rate, and the performance of HSS$_{10}$ hovered around 50\% at maximum. SSS failed to exceed the 30\% success rate, even with 50 retrieved candidates, and SSS$_1$ retrieved the optimal candidate in only around 10\% cases.

Finally, the number of retrieved candidates significantly impacts performance. Extending the list from 1 to 10 candidates yields a substantial improvement, but further extending it to 50 candidates offers only marginal additional gains, especially considering the increased workload for experts required to manually evaluate these candidates. For practical purposes, retrieving a single candidate is insufficient, delivering the correct structure in only about 20\% cases (HSS), depending on spectrum quality. 

These findings confirmed that current standard database search methods, while commonly used in practice, have limits in identifying unknown compounds.

\subsection*{SpecTUS results}
\begin{figure}
    \centering
    \includegraphics[width=0.8\textwidth]{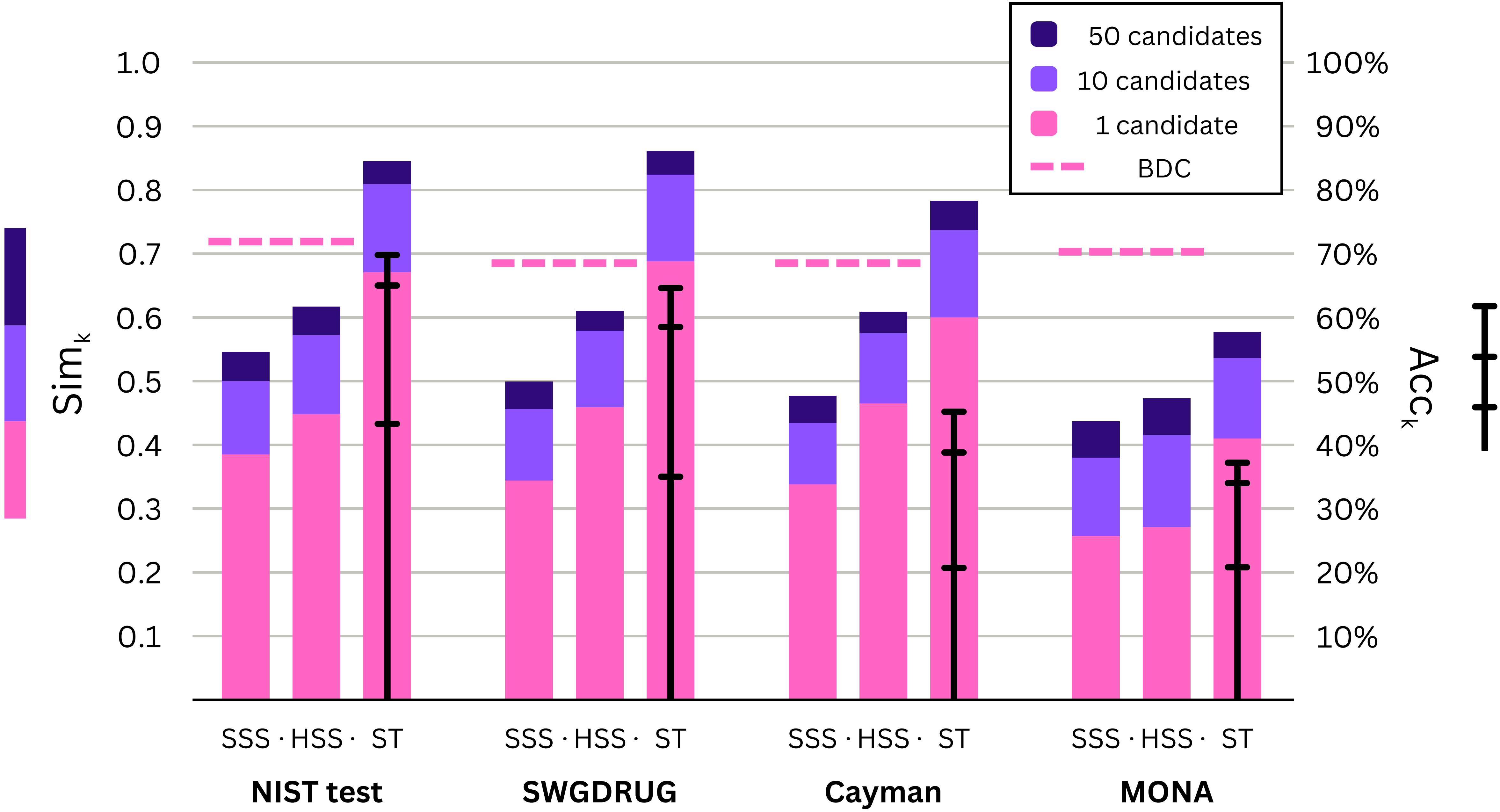}
    \caption{Comparison of average similarity and accuracy metrics across all tested methods, testing datasets, and three retrieval scenarios (1, 10, and 50 candidates). \textit{ST} represents SpecTUS, while the abbreviations of the baseline database search methods (\textit{SSS}, \textit{HSS}) are explained in the text -- Section \nameref{s:baselines}. Sim$_k$ is displayed by the color bars, similarity of the theoretical upper bound for database search methods (\textit{BDC}) is expressed as a pink dashed line. $\text{Acc}_k$ values for SpecTUS are shown in black lines; they are inherently zero for all the other methods. The three $\text{Acc}_k$ values in each column correspond to 1, 10, and 50 candidates, displayed from bottom to top.}
    \label{fig:acc_simk} 
\end{figure}

\begin{figure}
    \centering
    \includegraphics[width=0.48\textwidth]{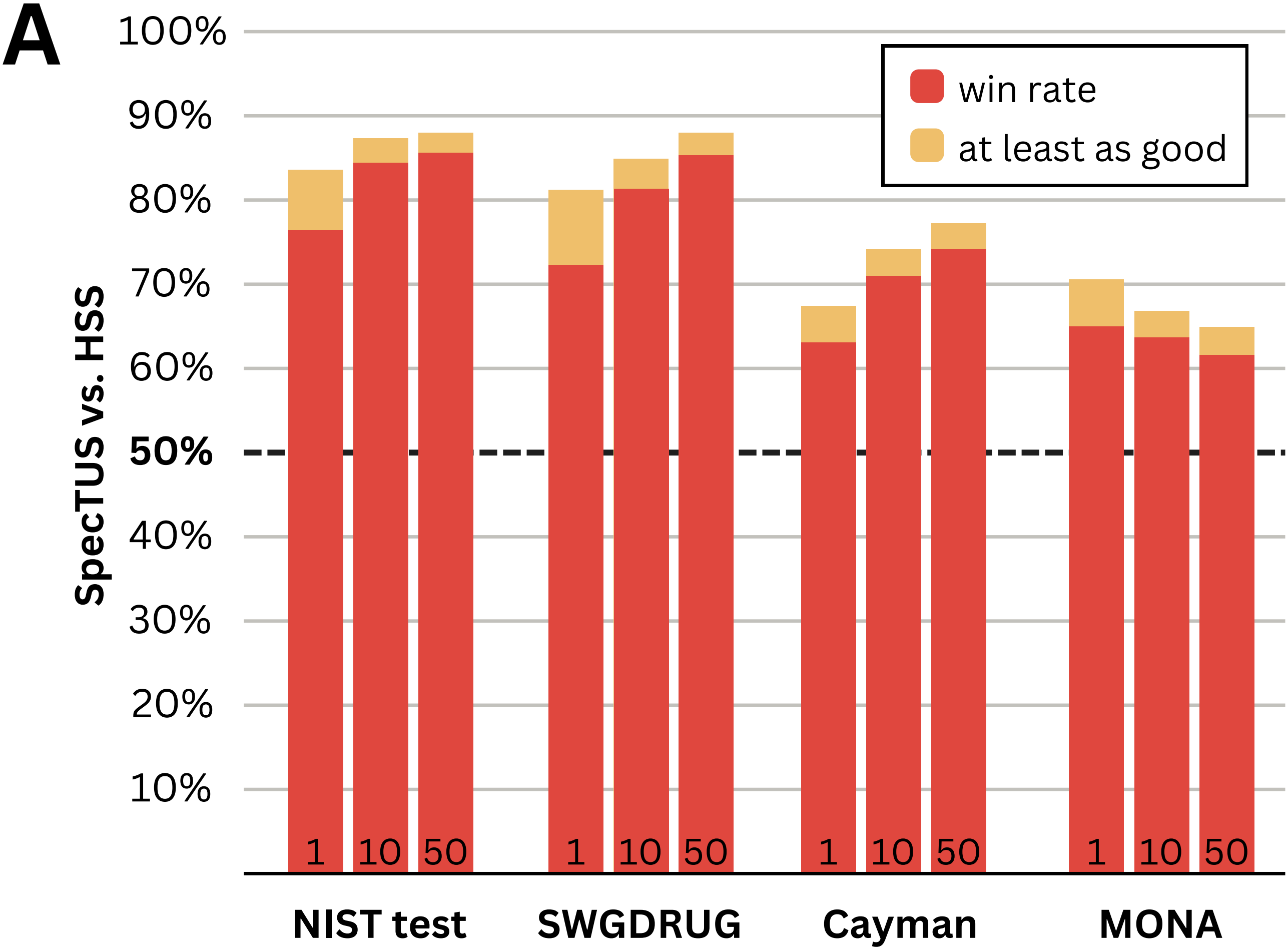}
    \hspace{0.02\textwidth}
    \includegraphics[width=0.48\textwidth]{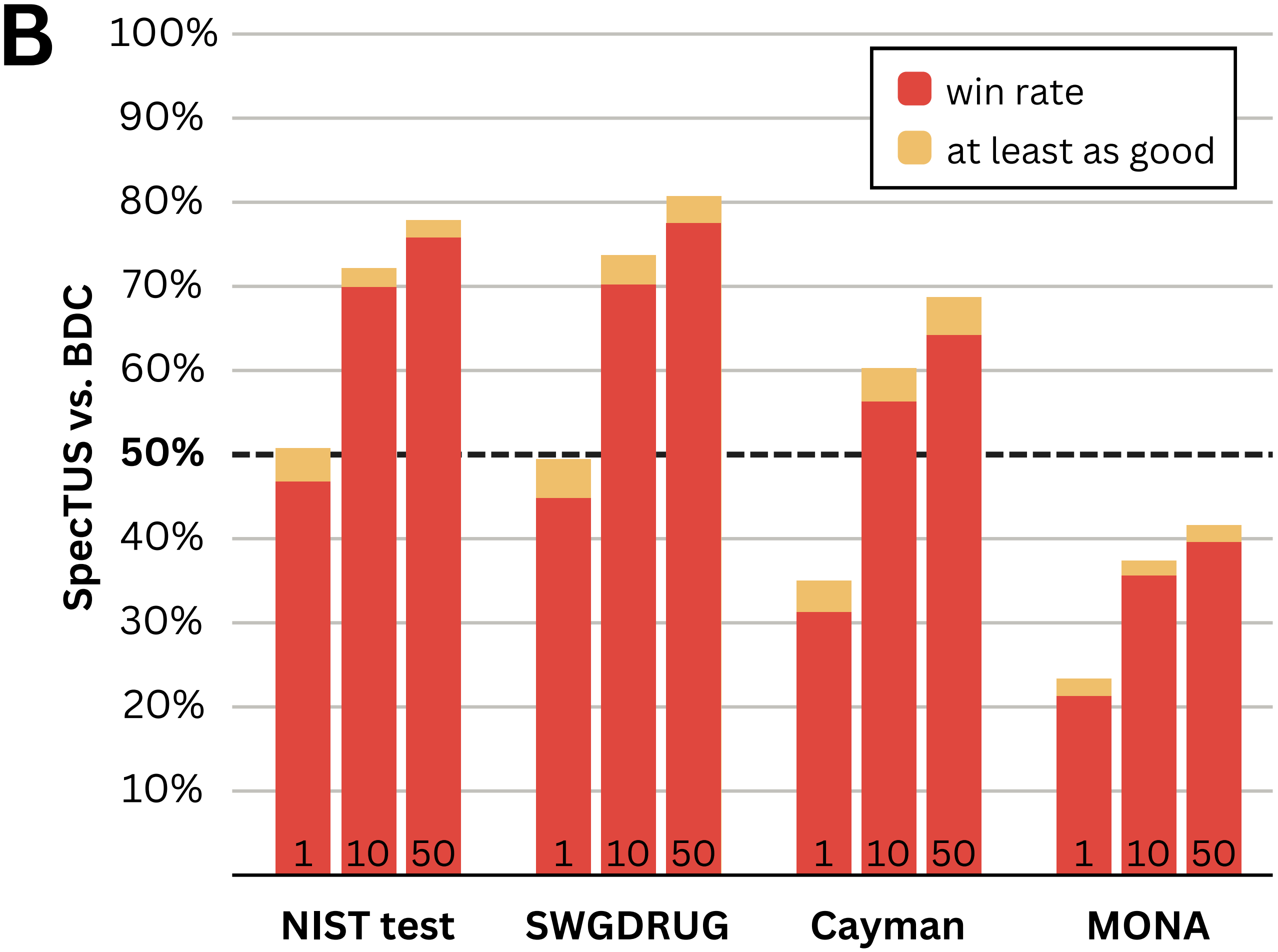}
    \caption{Comparison of  win rate and at-least-as-good rate of SpecTUS over database search method HSS (\textbf{A}) and a theoretical database search upper bound BDC (\textbf{B}) across all testing datasets, and three retrieval scenarios (1, 10, and 50 candidates). The diagram evaluates corresponding retrieval scenarios, such as SpecTUS$_{10}$ versus HSS$_{10}$ (Win$(\text{SpecTUS}_{10}, \text{HSS}_{10}$), providing a direct comparison of performance under identical conditions.}
    \label{fig:win_rate} 
\end{figure}

The essential results of SpecTUS comparison with traditional database search methods are shown in 
Fig.~\ref{fig:acc_simk} and Fig.~\ref{fig:win_rate}.
The tables with exact values supporting the figures are included in Sect.~\ref{app:spectus_results}.

SpecTUS' capability to generate candidates it has not encountered during training allows it to accurately reconstruct compounds that database search cannot identify precisely.
As shown in Figure~\ref{fig:acc_simk}, with just one candidate, the SpecTUS accuracy (finding the correct compound) Acc$_1$ is 43\% on the NIST test set. With 10 candidates, it reached the accuracy of 65\%. For the less curated datasets, the $\text{Acc}_k$ lowered, but even for the MONA dataset, SpecTUS reached $\text{Acc}_1$ of 21\% and $\text{Acc}_{10}$ of 34\%. The $\text{Acc}_{k}$ for any database search method in this scenario is inherently 0\%.

On the NIST test and SWGDRUG datasets, which contain thoroughly curated spectra, %
SpecTUS achieved an average top-1 similarity of 0.67 and 0.69, respectively,
outperforming HSS even with 50 candidates (62 for NIST and 0.61 for SWGDRUG), and it approached the theoretical best database performance that reached 0.72 for NIST and 0.69 for SWGDRUG.

In the ten-candidate scenario, SpecTUS improved further, achieving average similarities of 0.81 on NIST and 0.82 on SWGDRUG, surpassing even the theoretical limit of database search (BDC); this is an~evidence of SpecTUS' capability to generalize beyond 
the search among known structures.

For less curated datasets, such as Cayman Library and MONA, the performance gap between SpecTUS and database search methods has narrowed. However, even on these datasets, a single SpecTUS prediction matched the performance of 10 candidates retrieved by HSS.

The \textit{win} and \textit{at least as good} rates of SpecTUS compared to HSS and BDC across datasets (Figure~\ref{fig:win_rate} and Table~\ref{tab:win_alag_nist}) further illustrate the model's strengths. When compared to HSS (Figure~\ref{fig:win_rate}A), SpecTUS consistently performed better on more rigorously curated spectra (NIST test, SWGDRUG) and benefited more from generating additional candidates, except on the MONA dataset. Specifically, SpecTUS achieved win rates of 76–86\% on NIST, 72–85\% on SWGDRUG, 63–74\% on Cayman, and 62–65\% on MONA. When draws were included, the \textit{at least as good} rate increased further, reaching 80–90\% on the NIST test and SWGDRUG datasets. These findings demonstrate that SpecTUS is a far more reliable candidate generator for identifying novel molecules than standard HSS.

Figure~\ref{fig:win_rate}B shows SpecTUS' comparison with BDC, highlighting its ability to outperform even the best candidates in the database. On the NIST test and SWGDRUG datasets, a single SpecTUS candidate was strictly better than BDC in 47\% and 45\% cases, respectively.
However, with 10 candidates, SpecTUS outperformed BDC in 70\% of the cases for both the NIST test and SWGDRUG and in 56\% cases for the Cayman library. The \textit{at least as good} rate of SpecTUS$_{10}$ again raised the bar a bit higher, ranging between 72–74\% on the NIST test and SWGDRUG datasets. The \textit{win rate} on the MONA dataset ranged from 21-40\% which mirrors SpecTUS' decreased performance on less curated spectra.  Again, these results show SpecTUS' ability to effectively generalize over training data and to help mitigate the database coverage issue.

For the complete results comparing SpecTUS with database search methods, refer to the supplementary tables in \ref{tab:win_alag_nist}.

\subsection*{Examples}
\begin{figure}
    \centering
    \includegraphics[width=\textwidth]{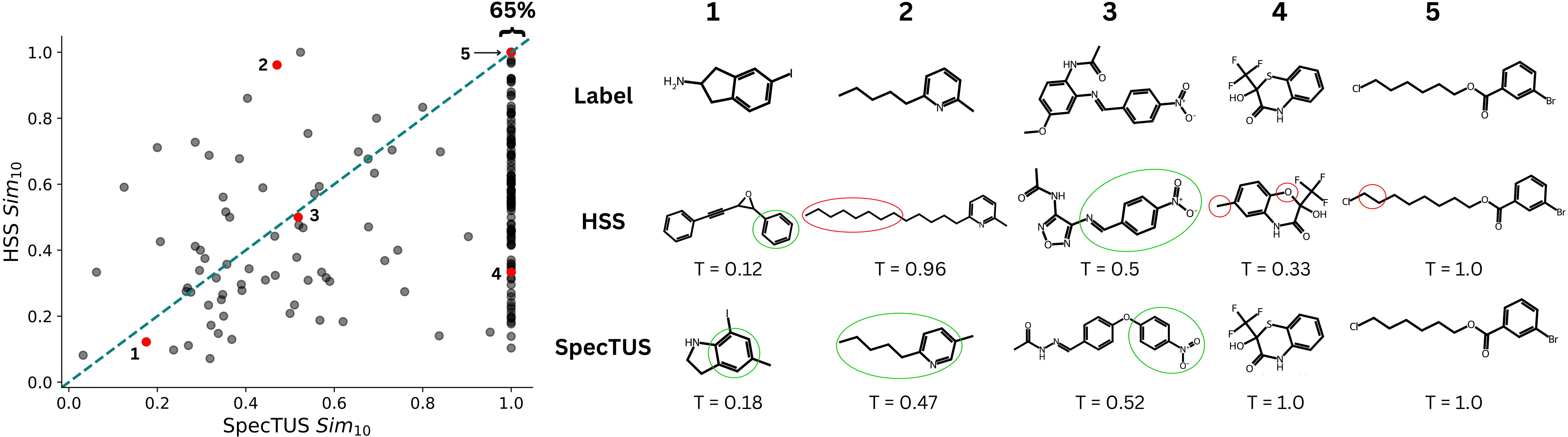}
    \caption{The scatterplot on the \textbf{left} illustrates the Sim$_{10}$ scores for 200 randomly sampled queries from the NIST test set, comparing SpecTUS and HSS predictions. Each point represents a single query, with its position determined by the Sim$_{10}$ scores of SpecTUS (x-axis) and HSS (y-axis). The dashed line indicates where both models achieved identical Sim$_{10}$ values. Notably, 65\% of the SpecTUS predictions reached a perfect Tanimoto similarity of 1.
    Highlighted in red are five specific examples, further detailed in the \textbf{table on the right}. For each example, the ground truth molecule (\textit{Label}) and the predictions from HSS and SpecTUS are shown, along with their Tanimoto similarity (T) to the label, computed using Morgan fingerprints. For faster comparison, correctly predicted regions are marked with \textit{green ellipses}, while errors are enclosed in \textit{red ellipses}. These examples were hand-picked to illustrate typical errors and highlight specific regions of the scatterplot.} 
    \label{fig:examples} 
\end{figure}

To better understand the predictive performance of SpecTUS and HSS, we analyzed their Sim$_{10}$ scores on 200 randomly sampled queries from the NIST test set. As shown in Figure~\ref{fig:examples}, the scatterplot compares the Sim$_{10}$ scores of both models, with the dashed diagonal indicating equal performance. Notably, SpecTUS achieved a perfect Tanimoto similarity score of 1 in 65\% of the queries, highlighting its ability to make accurate predictions more consistently than HSS. Additionally, five examples, highlighted in red on the scatterplot and detailed in the accompanying table, were handpicked to illustrate common error types that occur during predictions.

One frequent error involves extending or shortening repetitive chains, as seen in example 2 for HSS and example 5 for SpecTUS. Another typical issue is the addition of a correct functional group but on an incorrect atom in an~aromatic ring, such as in example 1 (SpecTUS) and example 2 (SpecTUS). A third type of error involves accurately decoding large molecular substructures but incorrectly connecting them, as illustrated in example 3 (SpecTUS). These examples serve to visually highlight typical challenges in molecular prediction tasks and provide a sense of how such errors look in practice.

Example 5 also highlights a limitation of the Tanimoto similarity metric when applied to molecular fingerprints. HSS prediction, in this case, achieves a perfect similarity score despite representing a different molecule from the ground truth. This issue arises particularly in long, repetitive chains, where structural differences may not impact the fingerprint representation. Tanimoto similarity is a useful and widely used measure, but its limitations should be considered, especially in cases where structural nuances are critical.

\subsection*{Inference speed}
The size of SpecTUS (354 million parameters) allows the model to be powerful enough to handle the task at hand yet lightweight enough to be meaningfully used in practical applications. To demonstrate its potential usage scenarios, we benchmarked SpecTUS inference on three hardware setups: a high-end GPU, a mid-range GPU, and a laptop-like CPU configuration. While exact prediction speeds depend on the specific hardware used, we provide a general overview in the Table~\ref{tab:inference_speed}.

Our benchmarks show that SpecTUS inference can run comfortably on a single CPU with 8 GB of RAM, processing a spectrum in approximately 8 seconds for a single generated candidate or 36 seconds for 10 candidates. Using even an affordable mid-range GPU significantly accelerates inference, enabling spectra to be analyzed within seconds or even fractions of a second.

\begin{table}[]
\caption{The time required to generate 1, 10, or 50 candidates for a single query was evaluated across three hardware settings. In the \textit{H100} setting, we used the top tier NVIDIA H100 80 GB graphics card, the \textit{RTX5000} setting used one mid-range NVIDIA Quadro RTX 5000 16GB GPU, and the \textit{CPU} setting uses one Xeon Gold 6130 CPU along with 8 GB of RAM.}
\begin{tabular}{l|l|l|l}
                                           & 1 candidate & 10 candidates & 50 candidates  \\ \hline
H100 & 0.2 s & 0.4 s & 1.9 s   \\
RTX5000& 0.5 s & 1.1 s & 7.6 s   \\
CPU & 8 s & 36 s & 450 s  \\ 
\end{tabular}
\label{tab:inference_speed}
\end{table}

\section*{Discussion}
SpecTUS addresses the problem of \textit{de novo} mass spectra reconstruction, which means a direct translation 
from the domain of mass spectra into the domain of molecular structures. It focuses on processing unseen 
spectra of unknown compounds, and it is the first model specifically designed to reconstruct molecular structures 
from low-resolution GC-EI-MS spectra without requiring additional supporting information like a precursor 
ion mass or molecular formula (on the contrary, SpecTUS predicts these even more accurately than recent methods~\cite{Moorthy2023}).

Our approach was inspired by existing \textit{de novo} MS/MS spectra reconstruction models, such as MSNovelist~\cite{msnovelist} or Spec2Mol~\cite{spec2mol}, and especially MassGenie~\cite{massgenie} which adopted various
concepts from natural language processing. In addition to adapting our model to a different type of spectra, 
we introduced several novel ideas, some inspired by neural machine translation. These include integrating 
multiple sources of synthetic spectra, innovative input encoding, and spectra source indication. Each new 
approach was experimentally evaluated to ensure the optimal configuration (see Section \ref{s:experiments}).

The SpecTUS's architecture is an encoder-decoder transformer with 354 million trainable parameters,
based on BART~\cite{Liu2020mBART}, developed for natural language processing originally. 
The model was pretrained on a~large set of synthetic spectra to 
develop a comprehensive understanding of the chemical space of small molecules, followed by finetuning on 
NIST 20, a~smaller high-quality set of experimental spectra. While this general pretraining-finetuning strategy 
has been employed in models like MassGenie and MSNovelist, SpecTUS introduces a unique approach to 
generating and integrating synthetic spectra, setting it apart from previous methods.

SpecTUS was evaluated on multiple test datasets, including a held-out subset of NIST and subsets of SWDRUG, Cayman, and MONA libraries. To align with the unknown compound identification scenario, the testing compounds were strictly disjoint from all compounds used for training both spectra generators and SpecTUS. Potential data leakage was avoided by excluding conflicting records based on canonicalized SMILES strings with stereoisomeric nuances removed, effectively ensuring no overlap between the training and testing sets. This rigorous data separation allowed us to confidently state that SpecTUS is capable of true generalization on unseen compounds rather than simply memorizing the training sets.  Across all the testing datasets, the model achieved molecule reconstruction accuracies ranging from 21--43\% (Acc$_{1}$) and 34--65\% (Acc$_{10}$)  as shown in Table~\ref{tab:final_model_comparison}.

On the other hand, SpecTUS, like many \emph{de novo} and extended database search approaches, cannot explicitly justify its specific outputs for a given input. It lacks an intermediate representation, such as peak annotations or fragmentation trees (cf.~\cite{HUFSKY2012fragTrees,wakoli2024gcms_id}), and it does not provide reference spectra for the candidate structures (typical in standard database searches). This limits the possibility of further manual correction of the predicted molecular structures. Additionally, the model’s poorer performance on the Cayman and MONA test sets -- compared to the more rigorously curated NIST or SWGDRUG libraries -- highlights its sensitivity to the quality of input spectra.

We compare SpecTUS with the traditional structure identification approach widely used by practitioners, which relies on searching reference spectra databases. This approach typically involves either exact peak matching (simple similarity search -- SSS) when the compound is suspected to be present in the database, or a more complex hybrid similarity search (HSS) to identify the closest structure when the compound is absent~\cite{hss_gcms_2017,hybrid_search}. 
Setting aside the combinatorial estimate of the number of all possible small molecules, which is
around $10^{60}$~\cite{reymond_chemical_spaces}, structural ``spectraless'' databases contain roughly $10^9$ unique molecules. In contrast, commercially available spectral libraries for database searching hold only a few hundred thousand spectra, significantly limiting the pool of potential matches for compound identification.
As we show on the NIST held-out testing set, in case the analyzed compound is missing from the reference library, the HSS is able to find the library's closest structure in only 19\% cases when retrieving the single best candidate. Practitioners typically refine these candidates through manual analysis and suggest structural corrections.  In the same setting, SpecTUS was able to offer a strictly better candidate than HSS in 76\% of the cases, and in 43\% it correctly identified the exact compound. 

The extended database search approaches attempt to address the issue of insufficient library coverage by enlarging it artificially.
DeepEI~\cite{deepei} achieves a top-1 accuracy (Acc$_1$) of 27.8\% (compared to SpecTUS, which ranges from 21\% to 43\%). However, we argue that its reference library of 170,000 fingerprints is far too small to adequately cover chemical space, making it insufficient for annotating truly unknown structures. Moreover, increasing the library size would inevitably decrease accuracy as a larger search space inherently leads to more false positives. DeepEI also relies on molecular weight information, which may not always be available, and is restricted to compounds containing only the 10 most common elements.
FastEI~\cite{fastei} reports an even higher Acc$_1$ of 36.7\% while expanding its reference library to 2.2 million synthetic spectra. However, we argue that this coverage remains inadequate, given the over 1 billion known compounds in ZINC, and it is still limited to just 11 elements. 
In contrast, SpecTUS was trained on 17.2 million spectra from 8.6 million compounds. Given its \textit{de novo} nature and demonstrated generalization ability, its predictions are not restricted to training spectra. Notably, as the training set size increased (Section \ref{s:experiments}), SpecTUS consistently improved in accuracy.


A straightforward comparison of SpecTUS with other \textit{de novo} approaches is challenging, as all existing models were designed and trained for MS/MS spectra. Although MS/MS methods suffer from lower output stability, they provide richer spectral information, allowing for the direct extraction of precursor ion mass. Notably, except for MassGenie, all previously published \textit{de novo} methods rely on precursor ion mass or molecular formula, which serve as valuable cues in the annotation process.
In terms of performance, MassGenie~\cite{massgenie}, which employs a similar transformer-based architecture as SpecTUS, reported an Acc$_{100}$ of 53\%. This was measured on an experimental testing set of 93 spectra filtered from the original 243 spectra in the CASMI 2017 competition using model-specific filters.
MSNovelist~\cite{msnovelist}  achieved an Acc$_{128}$ of 57\% on the CASMI 2016 competition dataset (127 spectra) and 45\% on the GNPS dataset (3863 spectra). However, MSNovelist relies on two streams of input -- spectral information encoded as a fingerprint and a precomputed molecular formula. When the authors removed the spectral information from the input, and the model was provided only the molecular formula, the reported Acc$_{128}$ dropped to 52\% and 31\% for CASMI 2016 and GNPS, respectively. This raises concerns about potential data leakage in the declared results.
Other \textit{de novo} models,
report significantly lower performance and are less comparable to SpecTUS.
MS2Mol~\cite{ms2mol} yields 21\% of ``close match'' accuracy, which is a far more relaxed metric.
Mass2SMILES~\cite{mass2smiles} appears to be rather premature, having less than 1\% of exact matches, and approx.~2\% of
close match accuracy.
In the most comparable scenario to existing \textit{de novo} methods, SpecTUS achieved an Acc$_{50}$ of 69\% on the NIST test set, which includes \numprint{28267} experimental spectra. 

To summarize, in this paper, we present SpecTUS, a novel end-to-end model for \textit{de novo} reconstruction of low-resolution GC-EI-MS spectra. Through a series of experiments, we demonstrate the impact of key design choices, offering transferable insights and good practices for future work. These include innovative input and output encoding strategies and the synergistic benefits of leveraging multiple synthetic spectra sources for pretraining. After identifying the optimal model architecture and training configuration, we rigorously compared SpecTUS to widely used database search techniques, showcasing its superior performance in terms of both identification accuracy and the molecular similarity of retrieved candidates.

We argue that while low-resolution GC-EI-MS spectra are less informationally rich compared to MS/MS spectra, SpecTUS achieves accuracy levels that are practically applicable in real-world scenarios.

Looking forward, we plan to explore how the increasing availability of high-resolution GC-MS data could further enhance molecular structure prediction. Additionally, we aim to investigate the effects of scaling up pretraining datasets to improve performance further.

\section*{Methods}\label{s:methods}

\begin{figure}
    \centering
    \includegraphics[width=.75\textwidth]{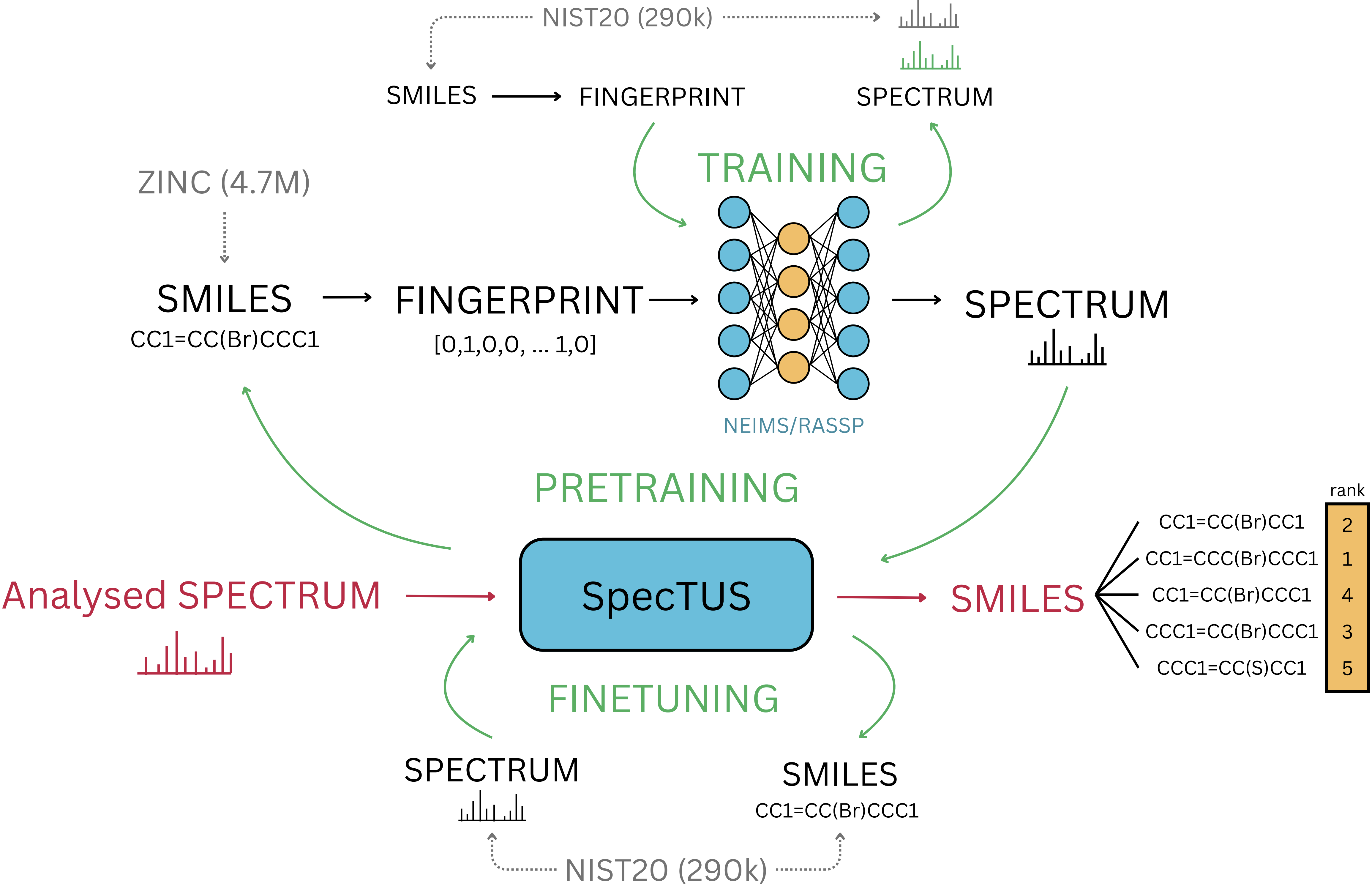}
    \caption{Overview of the SpecTUS method. The diagram illustrates relationships of all models (\textit{blue}), datasets (\textit{grey}) and training stages (\textit{green}) involved in constructing SpecTUS. It also highlights the final inference process (\textit{red}), showing how the model transitioned from training to producing ranked molecular predictions.}
    \label{fig:method} 
\end{figure}

Our model is built upon the standard Transformer
architecture~\cite{vaswani2017attention}, 
the BART~\cite{Liu2020mBART} in particular,
with certain design modifications and hyperparameters tuned specifically to optimize molecular structure annotation from mass spectra. The training process involved pretraining on large datasets of synthetic spectra generated using custom-trained NEIMS and RASSP models, followed by finetuning on the experimental spectra from the NIST 20 library.


\subsection*{Metrics}

Overall description of metrics used in SpecTUS evaluation and its comparison
with other models was given in Sect.~\nameref{s:results}. 
Here their detailed formal definitions are provided for completeness.

\textbf{Top-$k$ accuracy} ($\text{Acc}_k$) measures the percentage of predictions where the correct match is within the first $k$ retrieved candidates. It is formally defined as:
$$\text{Acc}_k(C_k, G):= \frac{100}{n} \sum_{i=1}^{n} \mathbbm{1}(G_i \in C_{ki})$$
where $k$ is the number of candidates considered, $n$ is the number of queries, $C_k$ is a list of $n$ predictions and each prediction $C_{ki}$ for query $i$ contains at most $k$ candidate molecules. $G$ is the list of ground truth molecules, one for each query, and $G_i$ is a single ground truth molecule for query $i$. $\mathbbm{1}$ is an indicator function that returns  1 if the condition inside is true and 0 otherwise.

\textbf{Average top-$k$ similarity} (Sim$_k$) quantifies how structurally similar, on average, the most accurate candidate among the first $k$ retrieved molecules is to the ground truth molecule. For each query, the similarity is calculated as the highest Tanimoto similarity between the Morgan fingerprints of the ground truth molecule and each of the first $k$ retrieved candidates. The similarities are then averaged over all queries in the testing set. Mathematically, the metric is defined as:
$$\text{Sim}_k(C_{k}, G) := \frac{1}{n} \sum_{i=1}^{n} \max\limits_{c \in C_{ki}}{\text{T}(\phi(c), \phi(G_i))}$$
where $T(f_1, f_2)$ denotes the Tanimoto similarity between two molecular fingerprints, and $\phi(m)$ is the Morgan fingerprint of molecule $m$.

Additionally, we derived two metrics to compare two sets of predictions, $C_k$ and $C'_k$, for a common set of queries: \textit{win rate} and \textit{at-least-as-good rate}. 

First, we had to define what it means for a prediction  $C_{ki}$ to be better than a prediction  $C'_{ki}$ given ground truth molecule $G_i$.  Tanimoto similarity on Morgan fingerprints is the standard approach to comparing molecules. However, because fingerprints are typically a lossy representation, situations can arise where $\text{T}(\phi(m_1), \phi(m_2))=1$ while $m_1 \neq m_2$. To address such ambiguities, we enhanced the ranking function with a control mechanism that checks for an exact match between candidates and the ground truth. This mechanism ensures that when both predictions achieve perfect similarity, the exact match determines superiority. The ranking function $R$ is then defined as follows:
$$ R(C_{ki}, C'_{ki}, G_i) = \begin{cases}
 & 1 \quad\text{ if } \max\limits_{c \in C_{ki}}{\text{T}(\phi(c), \phi(G_i))} > \max\limits_{c' \in C'_{ki}}{\text{T}(\phi(c'), \phi(G_i))} \\[1em]
 & 1 \quad \text{ if } G_i \in C_{ki} \land G_i \notin C'_{ki} \\[1em]
 & 0 \quad \text{ otherwise }
\end{cases}$$
\textbf{Win rate} calculates the percentage of cases where predictions $C_k$ are strictly better than predictions $C'_k$ according to the ranking function $R$. It is defined as:
$$\text{Win}(C_{k}, C'_{k}, G) := \frac{100}{n} \sum_{i=1}^{n} R(C_{ki}, C'_{ki}, G_i)$$
\textbf{At-least-as-good rate} extends the win rate by including cases where the two sets achieve equal top-$k$ similarity (draws). This metric completes the information about direct models' performance comparison by reflecting instances where predictions $C_k$ are either better than or equivalent to predictions $C'_k$:
$$\text{ALAG}(C_{k}, C'_{k}, G) := 100 - \text{Win}(C'_{k}, C_{k}, G)$$

\subsection*{Model}
SpecTUS utilizes an encoder-decoder architecture with 12 encoder blocks and 12 decoder blocks, 16 attention heads, an embedding size of 1024, and hidden layers of feed-forward blocks of size 4096. 

The model's input is a set of peaks represented by m/z and intensity values. The output is an autoregressively generated molecular structure representation encoded as a SMILES string. SpecTUS was trained on low-resolution GC-EI-MS spectra, and since integer m/z values already encode the relative position of peaks (unlike words/tokens in language models), we use the original positional encoding channel to encode intensity information. Thus, the model has three trainable sets of embeddings -- m/z values, binned intensities, and SMILES characters. The encoder receives a sum of embeddings representing m/z values and intensities -- one aggregated embedding for each peak in the spectrum; the decoder generation is based on a sequence of embeddings, each representing one SMILES character. 

Following the standard bounds for small molecules and NIST 20 library statistics, the model-specific thresholds for training and input data were set to a maximum of 300 peaks, an m/z limit of 500, and a maximum SMILES length of 100. Applying these thresholds resulted in excluding approximately 3\% of the NIST dataset. A detailed analysis of the impact of these filters is presented in Figure~\ref{fig:filters}.

In total, our model consists of 354 million trainable parameters.

\subsection*{Pretraining datasets}
\label{s:pretraining_datasets}
Pretraining on synthetic spectra is a core component of the SpecTUS model. For the early experiments, we collected 4.7 million compounds (\textit{synth1}) from the ZINC library and proved that pretraining greatly boosts the models' performance. For the last stage of SpecTUS development, we decided to step up the pretraining stage and double the size of the pretraining dataset by repeating the collection process once again. This time we gathered another 4.8 million compounds (\textit{synth2}).

To uniformly cover the known chemical space, we first scraped 1.8 billion SMILES strings from the ZINC20 library using the 2D-standard-annotated-druglike query. From this dataset, we extracted a random sample of 30 million noncorrupted SMILES strings shorter than 100 characters in each round. Further, we canonicalized, deduplicated, and stripped the SMILES of stereochemical information.
Importantly, we removed all NIST 20 compounds to prevent data leakage.
Filtering the 30-million sets to compounds RASSP can process (8 most common elements, number of possible subformulae)
reduces the dataset down to approx.~4.7 (\emph{synth1}) and 4.8 (\emph{synth2}) millions.
To evaluate the impact of RASSP and NEIMS respective spectrum prediction capabilities on SpecTUS performance,
we generated spectra for the same set of compounds by both models.
In total, the pretraining datasets comprises $9.4+9.6$ million spectra.

Lastly, we split each synthetic dataset into training, validation, and test sets using a $0.9:0.05:0.05$ ratio. The splitting process was random, but corresponding splits (training, validation, and test sets) for the NEIMS and RASSP-generated spectra contained the same compounds to avoid cross-leak. 
Both NEIMS- and RASSP-generated halves of both synthetic datasets (\textit{synth1} and \textit{synth2}) are available on the Hugging Face Hub for further research and replication.

\subsection*{Finetuning dataset}
\label{s:finetuning-dataset}
For the finetuning stage, we filtered all corrupted or nonannotated spectra from the low-resolution NIST 20 GC-EI-MS library, removed stereochemistry information, and canonicalized the SMILES strings. The dataset was split into training, validation, and test sets with an $0.8:0.1:0.1$ ratio, ensuring that duplicates were always placed in the same split. After applying model-specific filters, the final dataset included \numprint{224737} training spectra, \numprint{28177} validation spectra, and \numprint{28267} testing spectra.
The lists of compounds forming all three datasets are provided in the project's GitHub repository.

\subsection*{Preprocessing}
\label{s:preprocessing}
The preprocessing pipeline for spectra includes two key steps: rounding the m/z values to the nearest integer and logarithmically binning the relative intensity values into one of 30 bins using a logarithm base of 1.28 (details in Section \nameref{s:experiments_binning}). Additionally, each encoded SMILES sequence is enriched with a special token indicating the source of the spectra (\textit{<rassp>}, \textit{<neims>}, \textit{<nist>}). This strategy, inspired by multilingual language models~\cite{Conneau2019XLM,Liu2020mBART}, was designed to enable the model to adapt to the unique characteristics of spectra from different sources.

\subsection*{Pretraining}
SpecTUS was pretrained on both \textit{synth1} and \textit{synth2}, using a balanced 1:1 mixture of NEIMS-generated and RASSP-generated synthetic spectra. This pretraining stage provided the model with a broad understanding of the chemical space of small molecules, enhancing its ability to generalize effectively (as demonstrated in Section \nameref{s:experiments_mixing}). Pretraining was conducted with a batch size of 128 for \numprint{448000} updating steps, allowing the model to process each of the 17.2 million spectra approximately three times.

The entire pretraining process, including control evaluations on validation sets (single candidate) every \numprint{16000} steps, was completed in 58 hours using a single NVIDIA H100 GPU. To save time and computational resources during these evaluations, we used a random subset of \numprint{10000} NIST validation spectra and \numprint{30000} synthetic spectra.

During pretraining, the percentage of correctly reconstructed structures increased steadily but it remained relatively low at the end of the stage: 38\% for RASSP-generated spectra, 29\% for NEIMS-generated spectra, and 3\% for NIST spectra. However, 96\% of the generated SMILES strings (RASSP, NEIMS) were valid canonical molecules, with 91\% (RASSP), 78\% (NEIMS), and 14\% (NIST) having correct molecular formulas, though possibly incorrect structures. These results suggest that during the pretraining phase, the model successfully learned molecular structure rules and the relationship between atomic weight and m/z values, forming a good foundation for subsequent finetuning.

\subsection*{Finetuning}
In the second stage, SpecTUS was finetuned to enhance its ability to recognize and process real-world experimental data. The model was trained for \numprint{296000} steps on the NIST training set, allowing it to process each spectrum approximately 80 times.

The finetuning process, including control evaluations on validation sets every \numprint{21068} steps (approximately every 12 epochs), was completed in 28 hours on a single NVIDIA H100 GPU. To conserve time and computational resources, control evaluations used the full NIST validation set but were limited to a subset of \numprint{2000} synthetic spectra.

During finetuning, the model's ability to reconstruct structures from synthetic spectra dropped sharply to zero, with formula-matching accuracy decreasing to 19\% for NEIMS-generated spectra and 17\% for RASSP-generated spectra after just six epochs. However, validation results for NIST spectra skyrocketed above the convergence level of models trained exclusively on experimental spectra, as demonstrated in Section \nameref{s:experiments_mixing}.

\subsection*{Hyperparameter search}
\label{s:experiments}
One of the main contributions of this paper is a series of experiments that offer practical insights and best practices for developing machine-learning models, particularly transformer-based architectures, for spectrometry and chemistry. While some findings are technical, they provide valuable guidance to streamline the design and optimization of future models.

First, we investigated different methods for binning relative intensity values to effectively represent continuous data. We compared the performance of two molecular representations—SELFIES~\cite{Krenn_2020_SELFIES} and SMILES~\cite{Weininger1988SMILES}—in a generative modelling task and experimented with tokenization techniques for text-based molecular structures. Next, we examined the benefits of pretraining on synthetic data and identified best practices for dataset labeling and dataset mixing strategies. Lastly, we pushed the performance boundaries by leveraging larger datasets and extended training durations.

To optimize computational resources, we tailored our experimental approach. For the first two experiments, we skipped the pretraining phase and conducted only finetuning on the NIST dataset, using \numprint{74000} updating steps. For experiments focused on pretraining, we included a pretraining phase of \numprint{124000} steps, followed by a finetuning phase of \numprint{74000} steps. These run lengths were chosen to identify meaningful trends while maintaining manageable computational costs.

\subsubsection*{Experiment 1: Intensity binning}
\label{s:experiments_binning}
In the first experiment, we tried to find the optimal method for binning relative intensity values, enabling each value to be represented as a trainable embedding vector. We compared several variants of linear and logarithmic binning methods. For linear binning, we tested precision levels of 2, 3, and 4 decimal places, corresponding to 100, \numprint{1000}, and \numprint{10000} bins, respectively, each with its own trainable embedding. For logarithmic binning, we used the formula
$$n = \max(\lfloor\log_{b}(i)\rfloor+s),\ 0)$$
where $n$ is the assigned bin index to intensity value $i$, logarithm base $b$ changes the shape of the final bin distribution, and shift $s$ helps to set a particular number of bins for relative intensity values between 0 and 1. The $\max$ function ensures all the lowest intensities transformed to negative values fall into the bin 0. We evaluated four configurations, with the total number of bins ($s+1$) being 10, 21, 30, and 40. For each variant, we tuned the parameter $b$  on the NIST training set intensities to distribute the values into bins as evenly as possible (see the distributions in Section \ref{app:experiment_binning}). Binning beyond 40 bins was not practical, as the bin ranges became excessively narrow, leaving many bins empty and thus non-trainable.

Over-simplified binning approaches (linear binning with 2 decimal places and logarithmic binning with 10 bins) led to a 3–4\% drop in validation Acc$_1$ compared to the other models in the experiment. This performance decline was likely caused by significant information loss: for example, 47\% of the smallest intensity values were set to zero with 2-decimal-place linear binning. Similarly, 10-bin logarithmic binning lacked sufficient granularity for lower intensity values, as illustrated by histogram comparisons in Section \ref{app:experiment_binning}.

Higher-resolution binning methods—linear binning with 3 or 4 decimal places and logarithmic binning with 21, 30, or 40 bins—achieved comparable results, with validation performance varying within a 1\% margin. Among these, logarithmic binning with 30 bins emerged as the optimal choice. It provided a 0.9\% improvement in validation Acc$_1$ compared to 4-decimal-place linear binning while reducing the number of trainable parameters by 10 million, demonstrating its efficiency and validity as a binning strategy.

The tracked control evaluations for this experiment can be found in Figure~\ref{fig:exp_binning_acc}, and the comparison with the rest of the experiment runs is in Table~\ref{tab:experiments_comparison}.

\subsubsection*{Experiment 2: Molecular Representations and Tokenization}

In this experiment, we aimed to find the optimal molecular representation generated by the model's decoder. First, we evaluated two different structural representations for molecules, comparing SMILES and SELFIES. SELFIES, a token-based successor to the well-established SMILES representation, was specifically designed for generative neural networks, guaranteeing the validity of every sequence by construction.
Further, we explored enhancing classic character-level SMILES encoding through Byte Pair Encoding (BPE) tokenization~\cite{sennrich2016bpe}, which segments strings into frequent multi-character groups. This technique, widely adopted in early natural language generation (NLG) applications~\cite{Radford2018gpt1, Radford2019gpt2}, allows models to directly generate tokens with higher-level semantic meanings, potentially corresponding to molecular functional groups.

To investigate the impact of BPE tokenization, we trained four BPE tokenizers with varying vocabulary sizes on a random set of one million SMILES strings sampled from the synth dataset. The vocabulary size was indirectly controlled by adjusting the minimal substring frequency hyperparameter (mf) to values of 10, 100, 10K, and 10M. These settings yielded vocabulary sizes of \numprint{1286}, 780, 367, and 267 tokens, respectively, with the smallest vocabulary (mf10M $\rightarrow$ 256 byte tokens + 11 special tokens) corresponding to a character-level encoding. The relationship between vocabulary size and minimal substring frequency is visualized in Figure~\ref{fig:selected_tokenizers}.

Models using SMILES strings with BPE tokenization consistently outperformed the model trained on SELFIES, achieving improvements of 1.7–5.8\% in validation Acc$_1$. Interestingly, contrary to expectations based on NLG, larger BPE vocabularies did not lead to better performance. Instead, the character-level encoding (mf10M) outperformed all the BPE tokenizations by a margin of 3.5–4\% on validation Acc$_1$. The trend suggests not only that ``the smaller the vocabulary, the better the performance'' but also that even introducing relatively few aggregated tokens into the vocabulary can significantly degrade performance.

We currently lack a clear hypothesis to explain why SMILES outperformed SELFIES. However, based on the results of the models' comparison, we infer that the Transformer architecture effectively constructs higher-level molecular semantics internally within its decoder blocks. This suggests that keeping the representation simple, such as through character-level encoding, grants the model greater expressive freedom, ultimately leading to better performance.

The tracked control evaluations for the tokenization experiment can be found in Figure~\ref{fig:exp_tokenization_acc}, and the comparison with the rest of the experiment runs is in Table~\ref{tab:experiments_comparison}.

\subsubsection*{Experiment 3: Pretraining dataset mixing}
\label{s:experiments_mixing}
For pertaining experiments, we used the \textit{synth1} dataset (see Section \nameref{s:pretraining_datasets}) containing 4.7 million compounds or 9.4 million spectra generated by the NEIMS and RASSP models. To assess the impact of pretraining on the final model performance, we compared the 30-bin non-pretrained model from the last experiment to four models that were first pretrained on different mixtures of datasets and then finetuned on the NIST training set. The dataset mixing strategy involved random sampling from all datasets in the mixture without repetition, with the proportion of each dataset controlled by a weight parameter.

The pretraining experiments used four dataset combinations: RASSP-only, NEIMS-only, a 1:1 mix of NEIMS and RASSP (NEIMS:RASSP), and a mixture of NEIMS, RASSP, and a small fraction of NIST data in a 1:1:0.1 ratio (NEIMS:RASSP:NIST). The inclusion of NIST data in the last configuration was intended to accelerate the initial stage of finetuning and potentially keep this advantage throughout the training process.

Pretraining was conducted for a fixed number of \numprint{112000} steps, and with a batch size of 128, approximately 14.3 million examples were processed during pretraining. The number of epochs varied depending on the dataset mixture; for example, in the NEIMS:RASSP mix, each spectrum was seen approximately 1.7 times on average, while for the RASSP-only mixture, each spectrum was encountered about twice as many times. The performance of all models was compared on the NIST validation set after the finetuning phase. 

In the experiment, the models pretrained on synthetic datasets outperformed the finetuned-only model by 7–10\% on validation Acc$_1$, clearly exceeding the convergence level of the non-pretrained model. Among the pretrained models, the NEIMS-only model achieved 2\% better Acc$_1$ than the RASSP-only model. However, combining NEIMS and RASSP datasets in a 1:1 ratio led to the best performance, improving accuracy by an additional 1\% compared to NEIMS-only pretraining. Including a small fraction of experimental NIST data in the pretraining mix (NEIMS:RASSP:NIST) did not improve the final performance. 

From these results, we deduce that despite the questionable quality of synthetic spectra, having broader coverage of the chemical space during pretraining helps the model generalize better. Combining spectra from two sources is more effective than training on a single source for twice as many epochs. 

The tracked control evaluations for this experiment can be found in Figure~\ref{fig:exp_mixing_acc}, and the comparison with the rest of the experiment runs is in Table~\ref{tab:experiments_comparison}.

\subsubsection*{Experiment 4: Source indication}

After identifying the optimal pretraining strategy, we evaluated whether the model benefits from including source indication via a special token preceding the generated SMILES sequence (see Section \nameref{s:preprocessing}). To test this, we compared the RASSP:NEIMS pretrained and NIST finetuned model from the previous experiment, which used distinct source tokens to indicate the origin of each spectrum, with an otherwise identical setting that utilized a single generic source token for all datasets.

The experiment revealed that the source indication had no substantial impact on performance. It did not affect control validations during pretraining, and at the end of finetuning, the model with a single source token performed 0.2\% better on Acc$_1$ than the one with three distinct source tokens. Despite this, we retained the source indication mechanism in the final model to allow for potential use in future finetuning experiments on smaller custom datasets. Users can decide whether to explore this feature further or use the default <nist> token on their data.

The tracked control evaluations for this experiment can be found in Figure~\ref{fig:exp_source_ind_acc} and the comparison with the rest of the experiments runs is in Table~\ref{tab:experiments_comparison}.

\subsubsection*{Experiment 5: Training length and dataset size}
In the final experiment, we assessed the impact of extended training duration and a larger dataset on the model's performance. Using the best-performing model from Experiment 3 as the baseline -- pretrained on \textit{synth1} NEIMS:RASSP balanced mix (4.2 million training compounds) for \numprint{112000} steps and finetuned on the NIST experimental spectra for \numprint{74000} steps -- we explored various combinations of extended pretraining and finetuning lengths, as well as larger training datasets. All other hyperparameters, including those related to binning, tokenization, and dataset mixing, were kept consistent with the findings of Experiments 1-3.

To simplify comparisons, we used a naming convention: the number of pretraining compounds, pretraining steps, and finetuning steps (e.g., 4.2M\_112k\_74k for the baseline model). The tested combinations included (1) doubling the pretraining duration to 224k steps while keeping finetuning at 74k steps (4.2M\_224k\_74k), (2) doubling both pretraining and finetuning steps (4.2M\_224k\_148k), (3) pretraining on the combined \textit{synth1} and \textit{synth2} datasets with prolonged pretraining and finetuning (8.6M\_224k\_148k), and (4) a full-scale training setting with both datasets and quadrupled pretraining and finetuning to maximize convergence (8.6M\_448k\_296k).

Doubling the pretraining stage alone to \numprint{224000} steps (4.2M\_224k\_74k) improved Acc$_1$ by 1.2\%. Further doubling the finetuning stage (4.2M\_224k\_148k) increased Acc$_1$ by another 0.7\%. Expanding the dataset size to include both \textit{synth1} and \textit{synth2} (8.6M\_224k\_148k) yielded a 0.9\% gain in Acc$_1$. These isolated improvements indicate that extending training duration and increasing dataset size independently benefit the model. While we lack the analysis of all combinations to draw definitive conclusions, the results suggest that scaling pretraining, finetuning, and dataset size contribute similarly to performance gains. Still, among the tested approaches, expanding the dataset size was the most effective option, as it required no additional computing power. 

The best results were achieved by combining all three strategies. The final configuration (8.6M\_448k\_296k) improved Acc$_1$ by 2.6\% compared to the intermediate 8.6M\_224k\_148k setting and by 5.4\% compared to the baseline 4.2M\_112k\_74k model. This fully converged SpecTUS model represents the highest-performing configuration across all experiments.

The tracked control evaluations for this experiment can be found in Figure~\ref{fig:exp_train_len_acc}, and the comparison with the rest of the experiment runs is in Table~\ref{tab:experiments_comparison}.

\backmatter

\bmhead{Supplementary information}

\begin{itemize}
\item 
For a~quick, zero-install demonstration, SpecTUS inference (with limited throughput) is exposed via REST API at our cluster,
accessible with a~Binder-ready Jupyter notebook at \url{https://github.com/ljocha/spectus-demo}.
\item
Complete source code of SpecTUS is available on GitHub: \url{https://github.com/hejjack/SpecTUS}.
\item 
The whole process of building the model
from scratch or fine-tuning the pretrained model, and reproducing the test results of this paper,
is implemented in a~series of Jupyter notebooks included in the repository above.
\item
Pretrained model and synthetic datasets are published at \url{https://huggingface.co/MS-ML}.
\item 
Further tables and figures to support reasoning in the main paper are provided separately.
\end{itemize}

\bmhead{Acknowledgements}
%
%
This work was supported by the Ministry of Education, Youth and Sports of the Czech Republic through the projects \emph{e-Infrastruktura CZ} (No.~LM2023054) and 
\emph{RECETOX Research Infrastructure} (No.~LM2023069), 
and European Union’s Horizon 2020 research and innovation program under grant agreement No.~857560 (\emph{CETOCOEN Excellence}). This publication reflects only the author's view, and the European Commission is not responsible for any use that may be made of the information it contains.
Computational resources were provided by the \emph{e-INFRA CZ} project (No.~90254),
supported by the Ministry of Education, Youth and Sports of the Czech Republic.

\bibliography{sn-bibliography} 

\newpage
\begin{appendices}

\section{Results}

\subsection{Baseline results}
\label{app:baseline_results}
\input{tables/db_searches}

\subsection{SpecTUS results}
\label{app:spectus_results}
\input{tables/final_model_eval}
\input{tables/win_alag}

\clearpage\newpage
\section{Methods}\label{secA1}

\begin{figure}[h]
    \centering
    \includegraphics[width=0.49\textwidth]{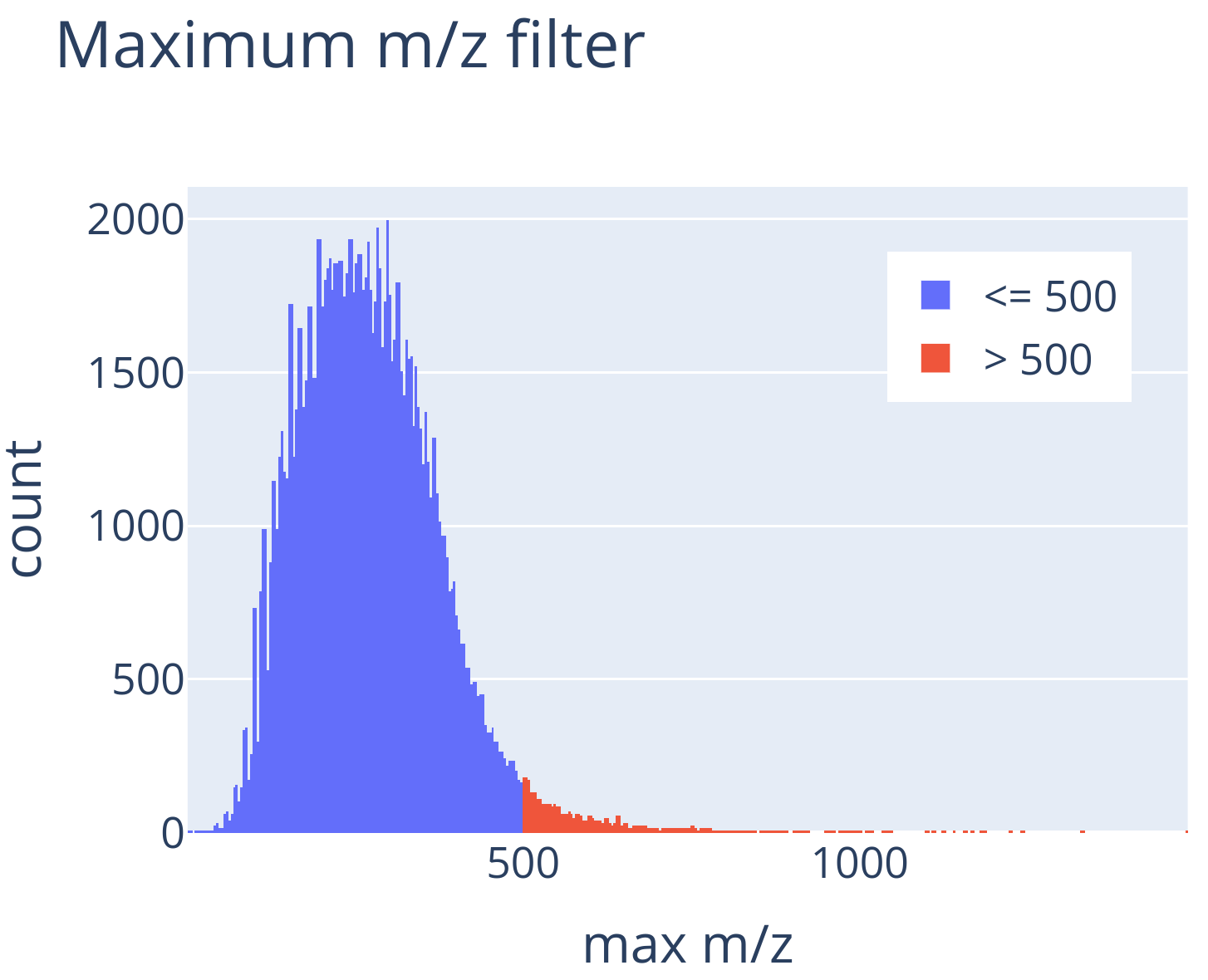}
    \includegraphics[width=0.49\textwidth]{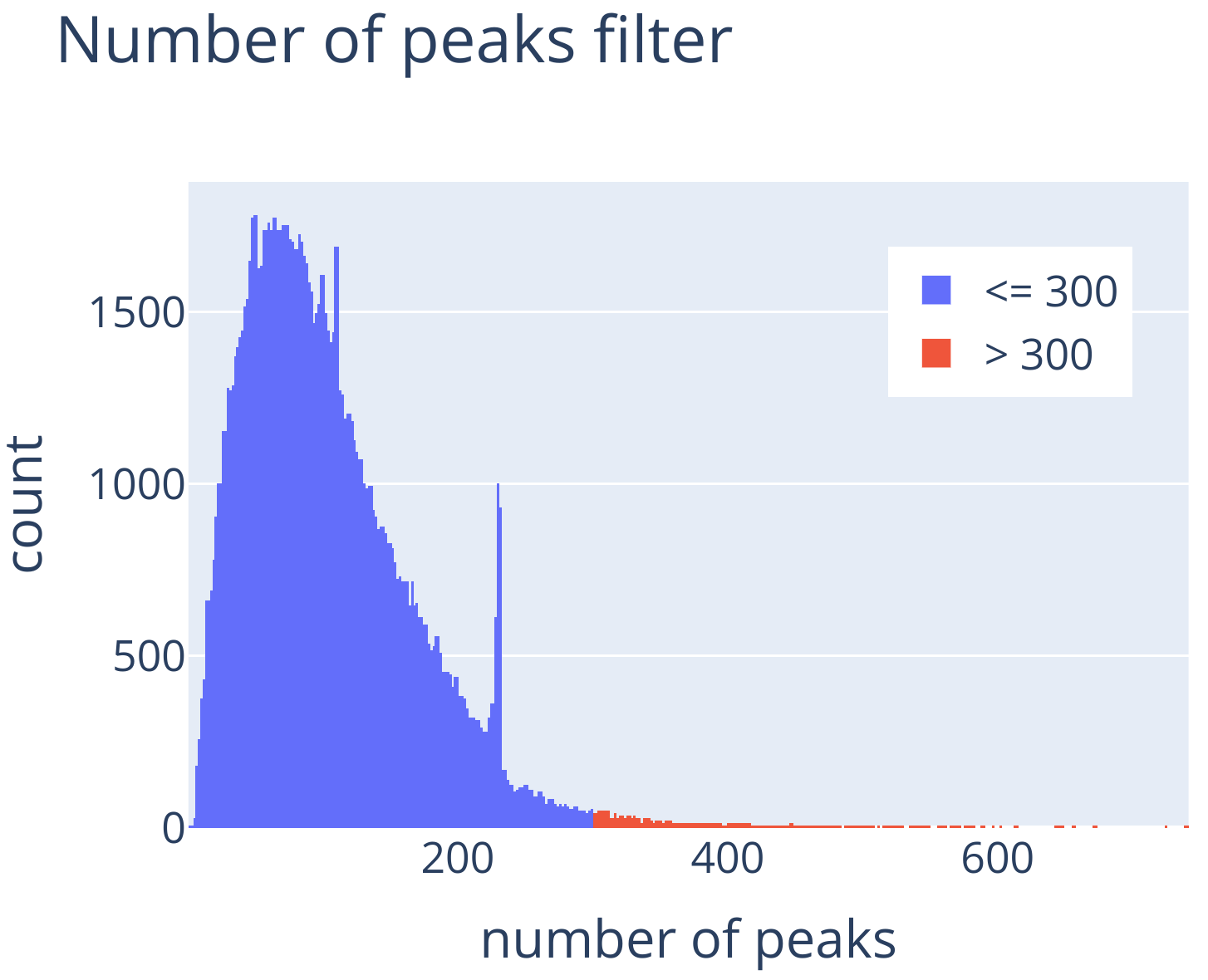}
    \vspace{5mm}    
    \includegraphics[width=0.49\textwidth]{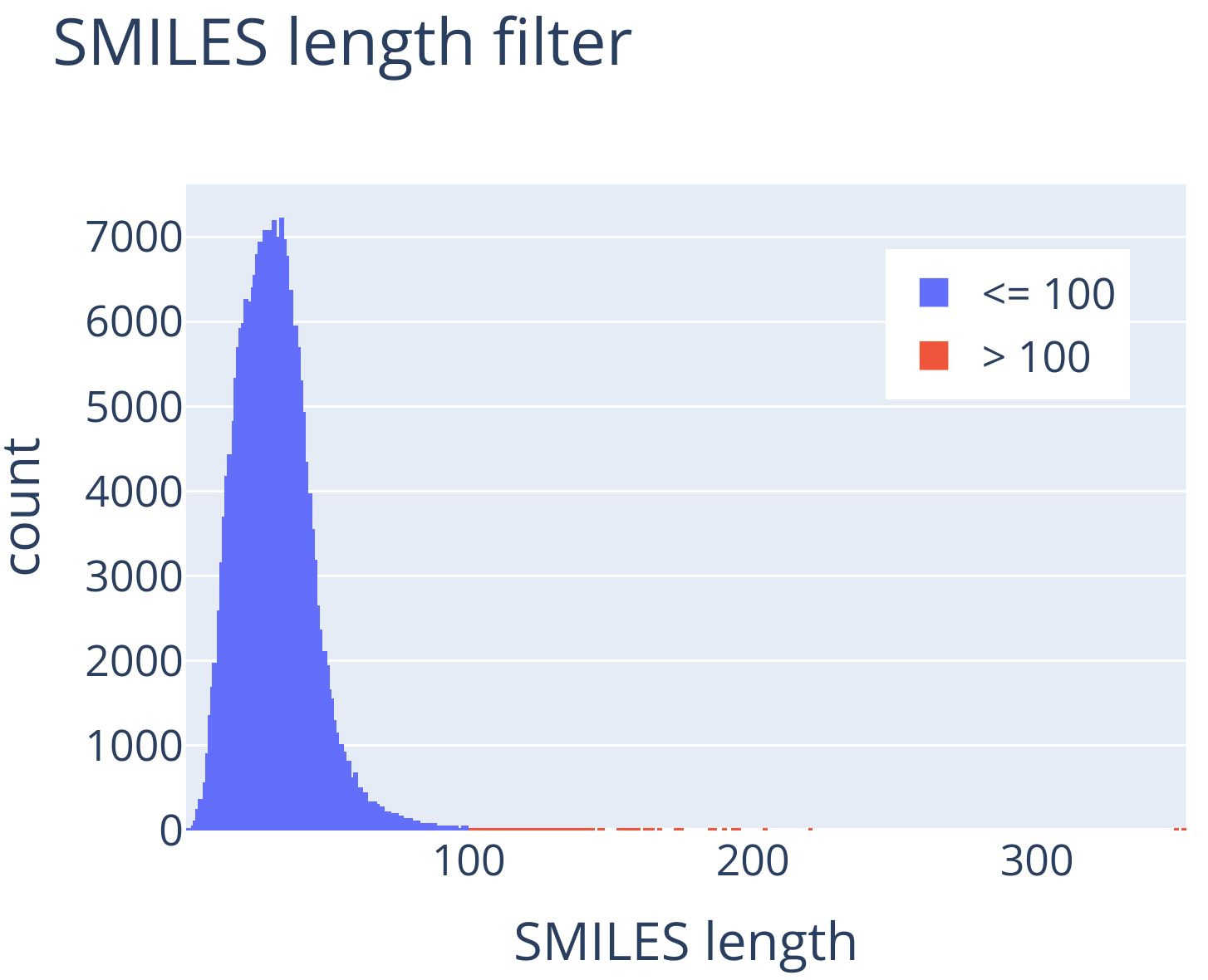}
    \caption{Frequency plots of the three data filtering criteria, measured on the NIST train set.}
    \label{fig:filters}
\end{figure}

\clearpage\newpage
\section{Experiments}

\input{tables/all_experiments}

\subsection{Experiment 1: Intensity binning}
\label{app:experiment_binning}

\begin{figure}[h]
    \centering
    \includegraphics[width=\textwidth]{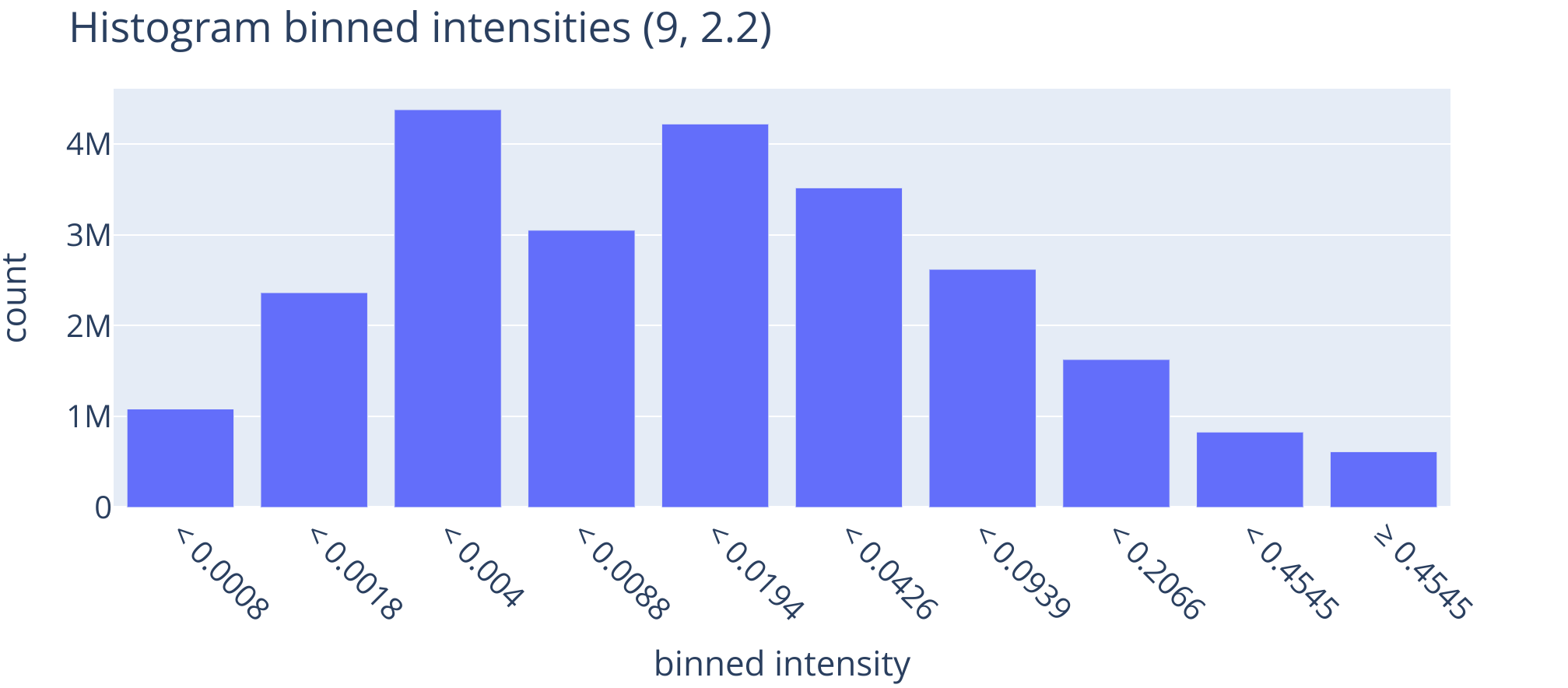}
    
    \caption{Bin frequency plot after logarithmic binning with a log base of 2.2 and a shift of 9, producing 10 bins. The distribution is calculated on our train split of the NIST dataset.}
    \label{fig:histo10bin}
\end{figure}

\begin{figure}[h]
    \centering
    \includegraphics[width=\textwidth]{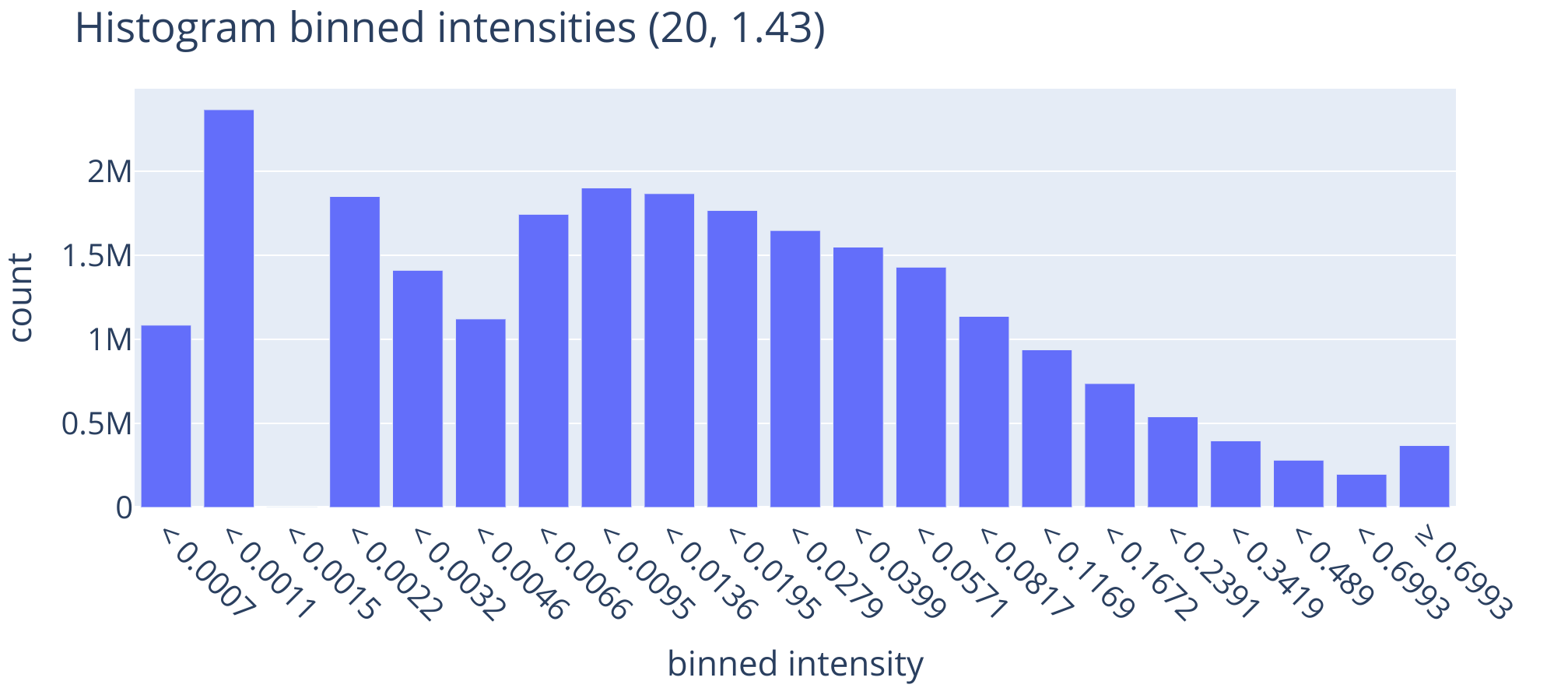}
    
    \caption{Bin frequency plot after logarithmic binning with a log base of 1.43 and a shift of 20, producing 21 bins. The distribution is calculated on our train split of the NIST dataset.}
    \label{fig:histo21bin}
\end{figure}

\begin{figure}[h]
    \centering
    \includegraphics[width=\textwidth]{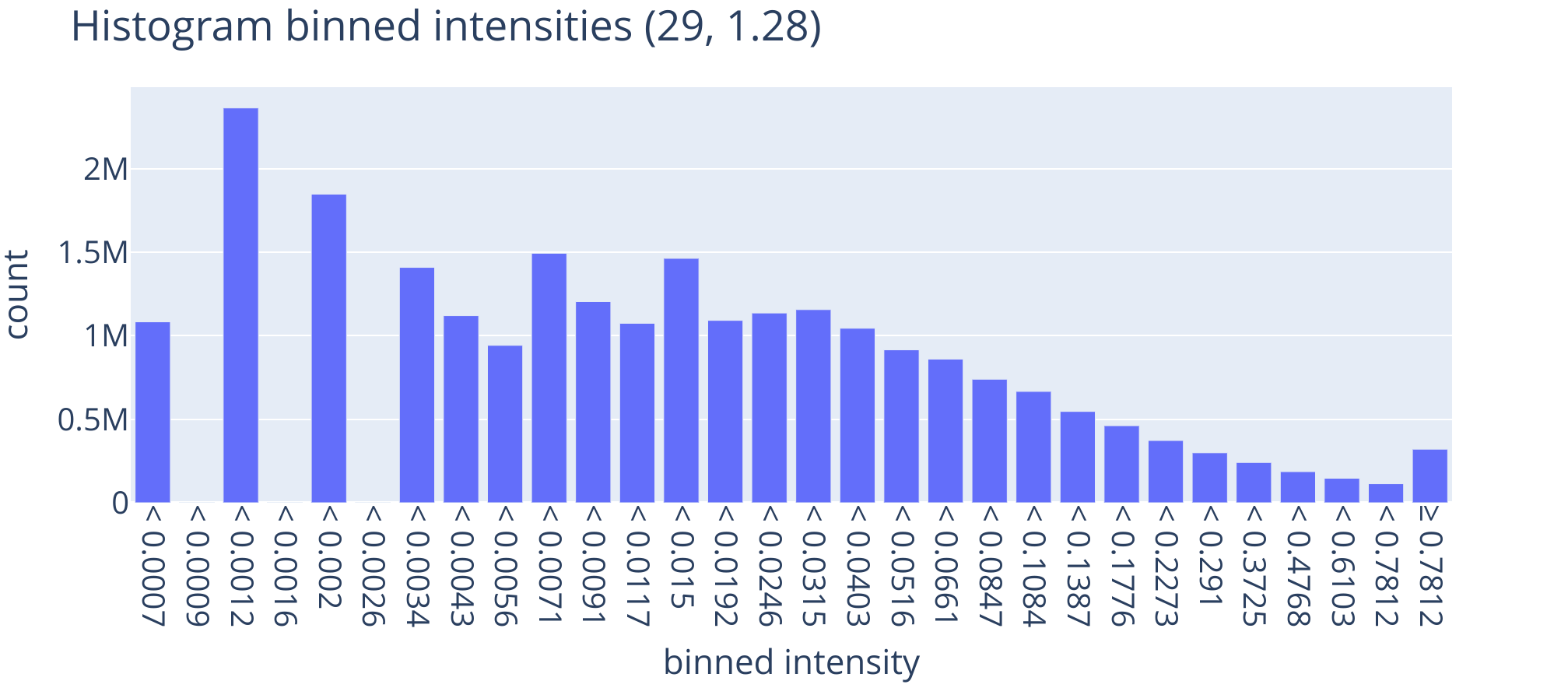}
    
    \caption{Bin frequency plot after logarithmic binning with a log base of 1.28 and a shift of 29, producing 30 bins. The distribution is calculated on our train split of the NIST dataset.}
    \label{fig:histo30bin}
\end{figure}

\begin{figure}[h]
    \centering
    \includegraphics[width=\textwidth]{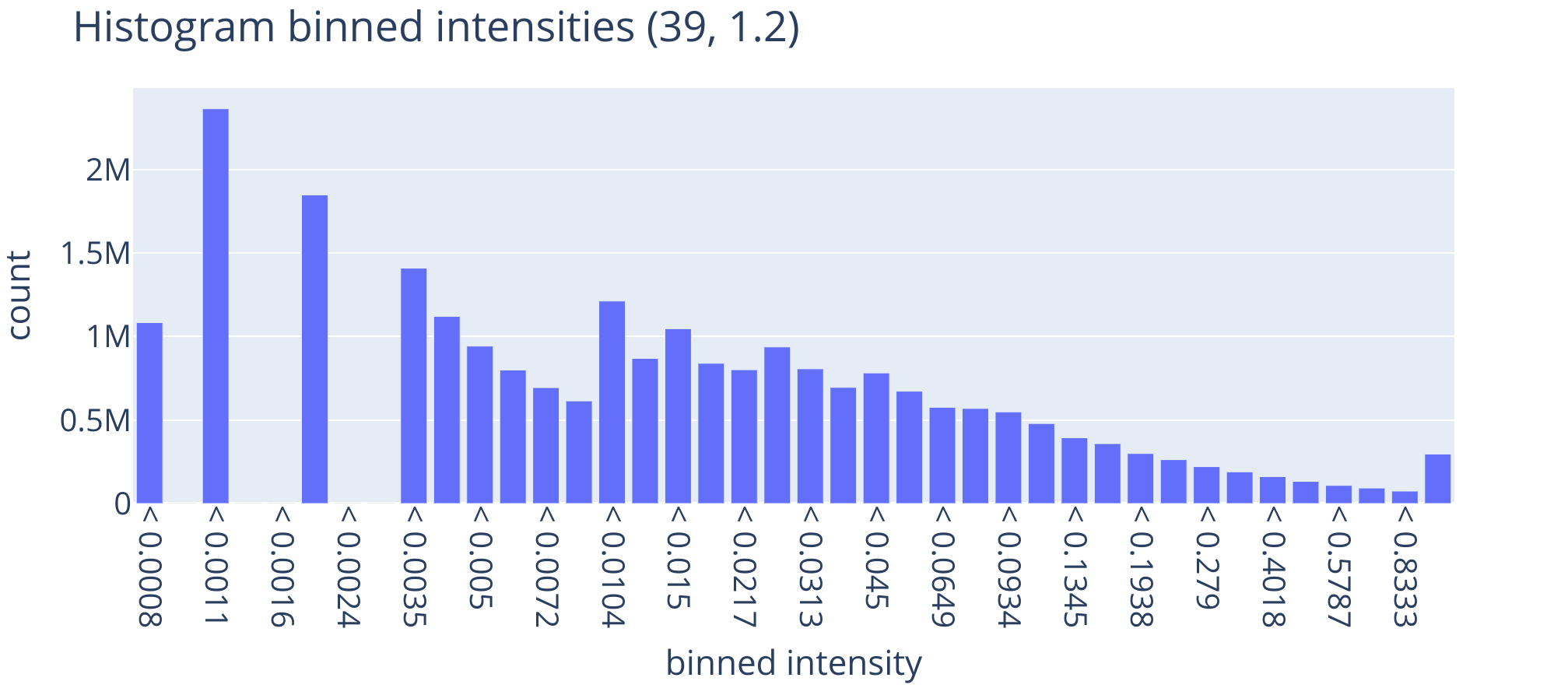}
    
    \caption{Bin frequency plot after logarithmic binning with a log base of 1.2 and a shift of 39, producing 40 bins. The distribution is calculated on our train split of the NIST dataset.}
    \label{fig:histo40bin}
\end{figure}

\begin{figure}[h]
    \centering
    \includegraphics[width=0.8\textwidth]{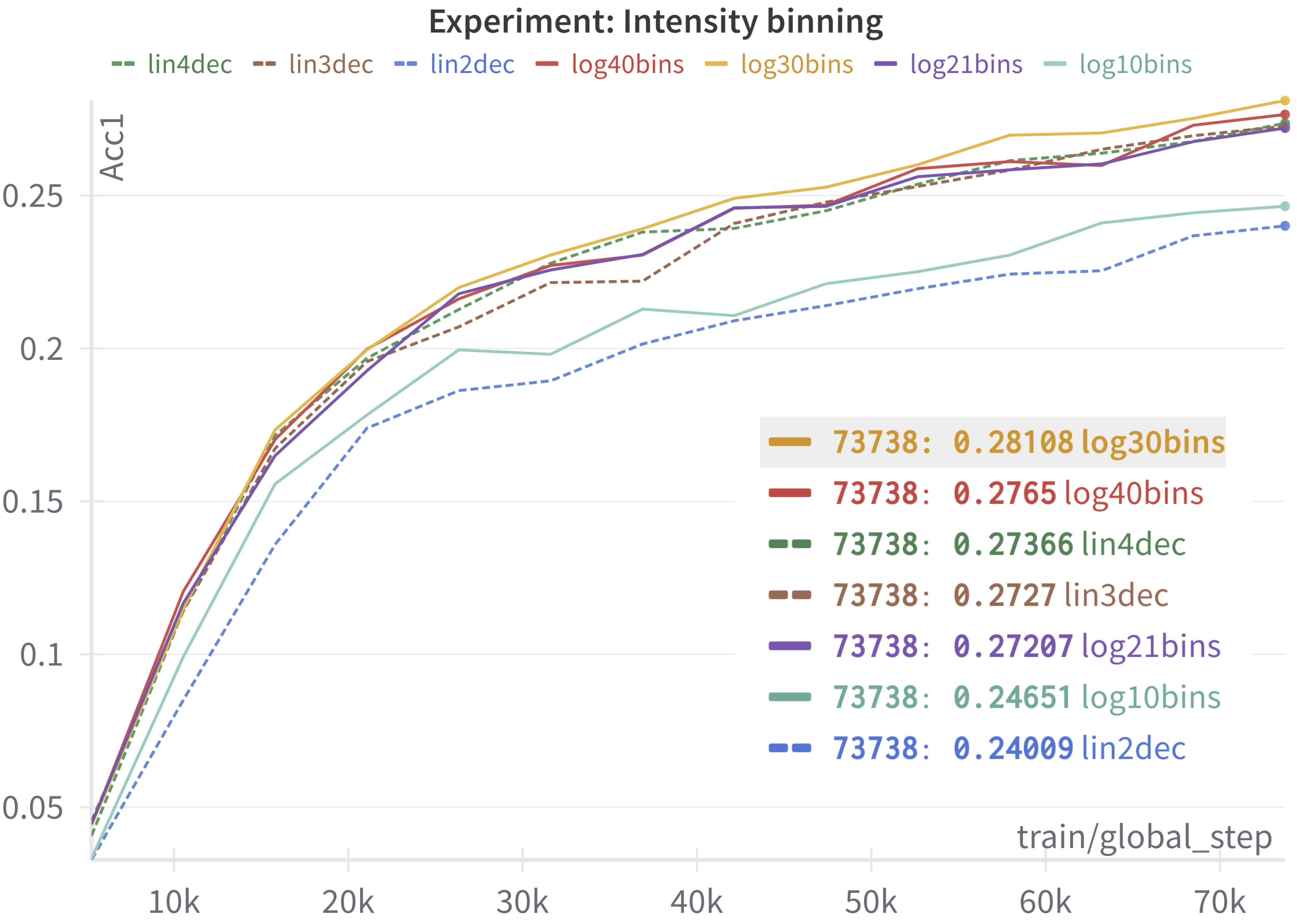}
    
    \caption{Tracked control evaluations of the \textbf{intensity binning} experiment. 
    Displays the dependency of Acc$_1$ on global step measured on the full NIST validation set. The table shows the final Acc$_1$ values.}
    \label{fig:exp_binning_acc}
\end{figure}


\clearpage\newpage
\subsection{Experiment 2: Molecular Representations and Tokenization}
\label{app:experiment_tokenization}

\begin{figure}[h]
    \centering
    \includegraphics[width=\textwidth]{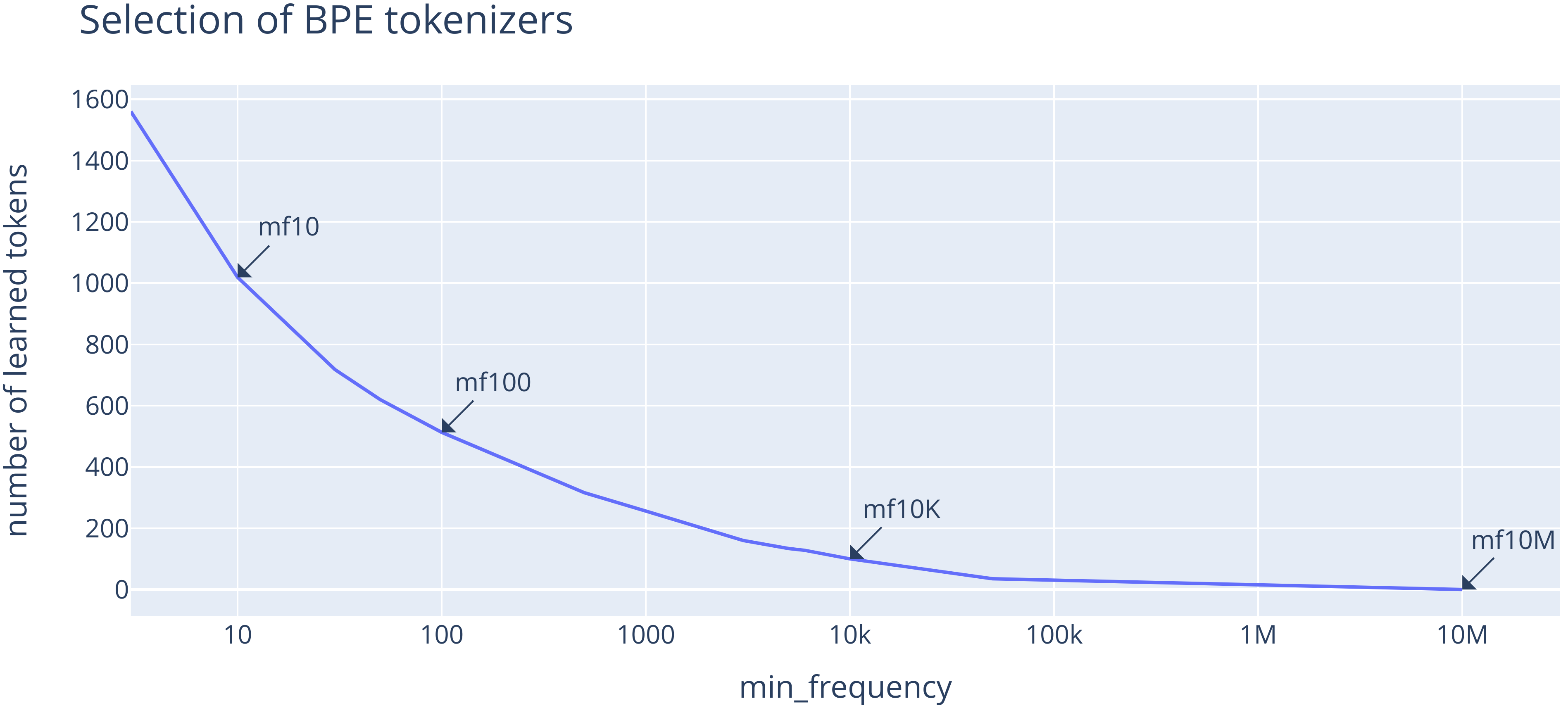}
    
    \caption{Dependence of the number of learned tokens on the minimal frequency parameter. The highlighted points denote the tokenizers that we selected for examination. The x-axis is in logarithmic scale.}
    \label{fig:selected_tokenizers}
\end{figure}

\begin{figure}[h]
    \centering
    \includegraphics[width=0.8\textwidth]{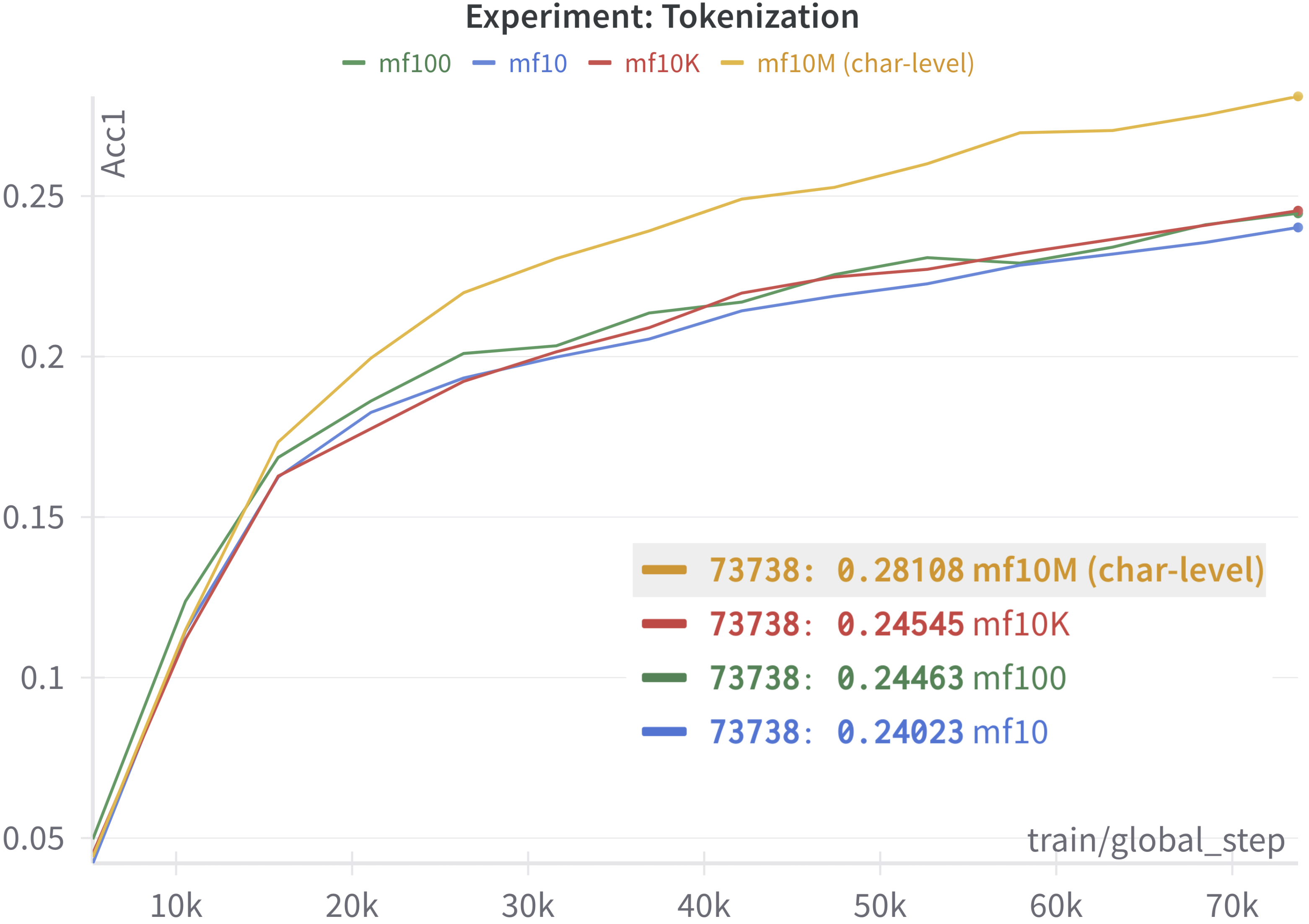}
    
    \caption{Tracked control evaluations of the \textbf{tokenization} experiment. 
    Displays the dependency of Acc$_1$ on global step measured on the full NIST validation set. The table shows the final Acc$_1$ values.}
    \label{fig:exp_tokenization_acc}
\end{figure}

\clearpage\newpage
\subsection{Experiment 3: Pretraining dataset mixing}
\label{app:experiment_mixing}

\begin{figure}[h]
    \centering
    \includegraphics[width=0.8\textwidth]{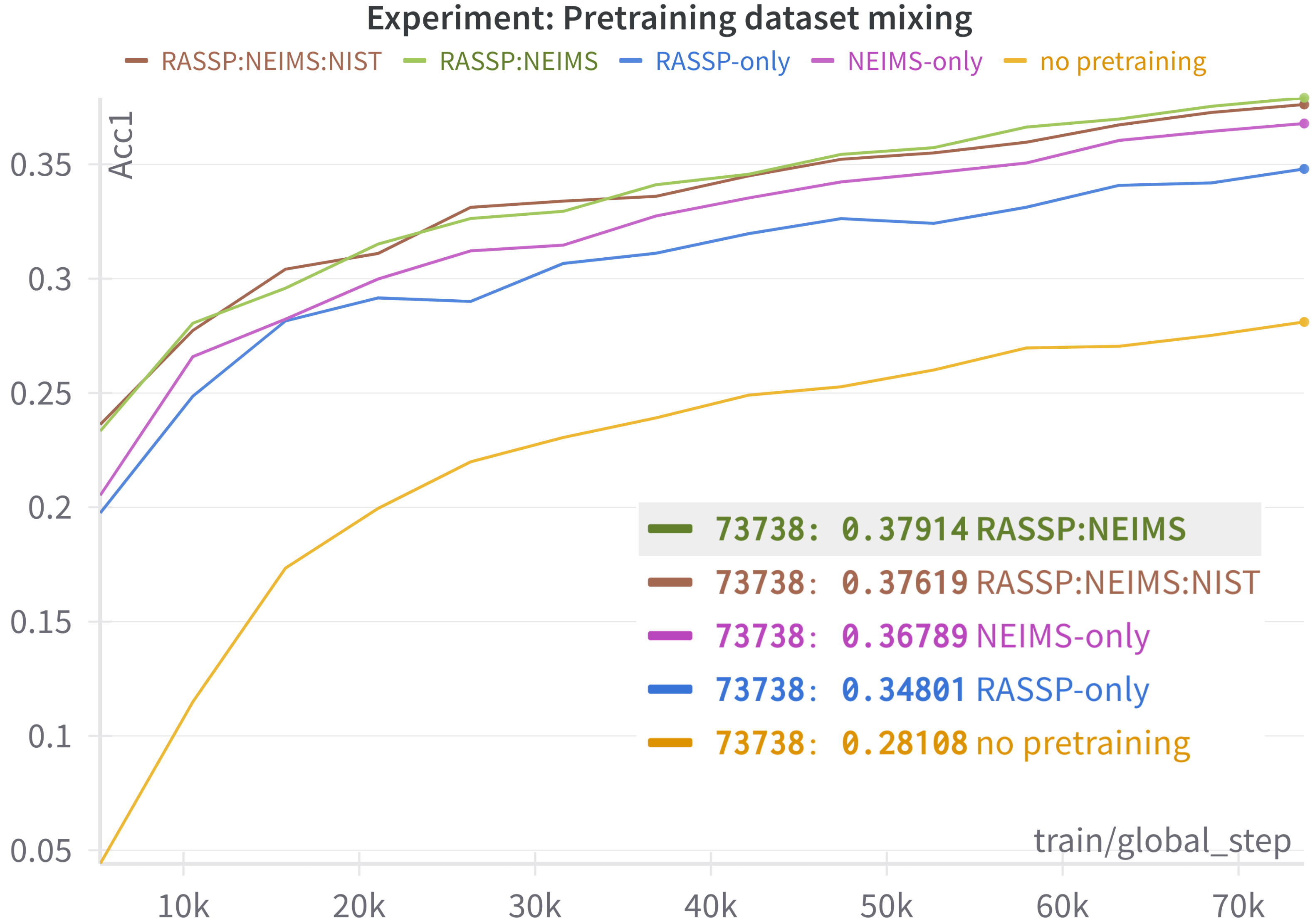}
    
    \caption{Tracked control evaluations of the \textbf{pretraining dataset mixing} experiment. 
    Displays the dependency of Acc$_1$ on global step measured on the full NIST validation set. The table shows the final Acc$_1$ values.}
    \label{fig:exp_mixing_acc}
\end{figure}

\clearpage\newpage
\subsection{Experiment 4: Source indication}
\label{app:experiment_source_ind}

\begin{figure}[h]
    \centering
    \includegraphics[width=0.8\textwidth]{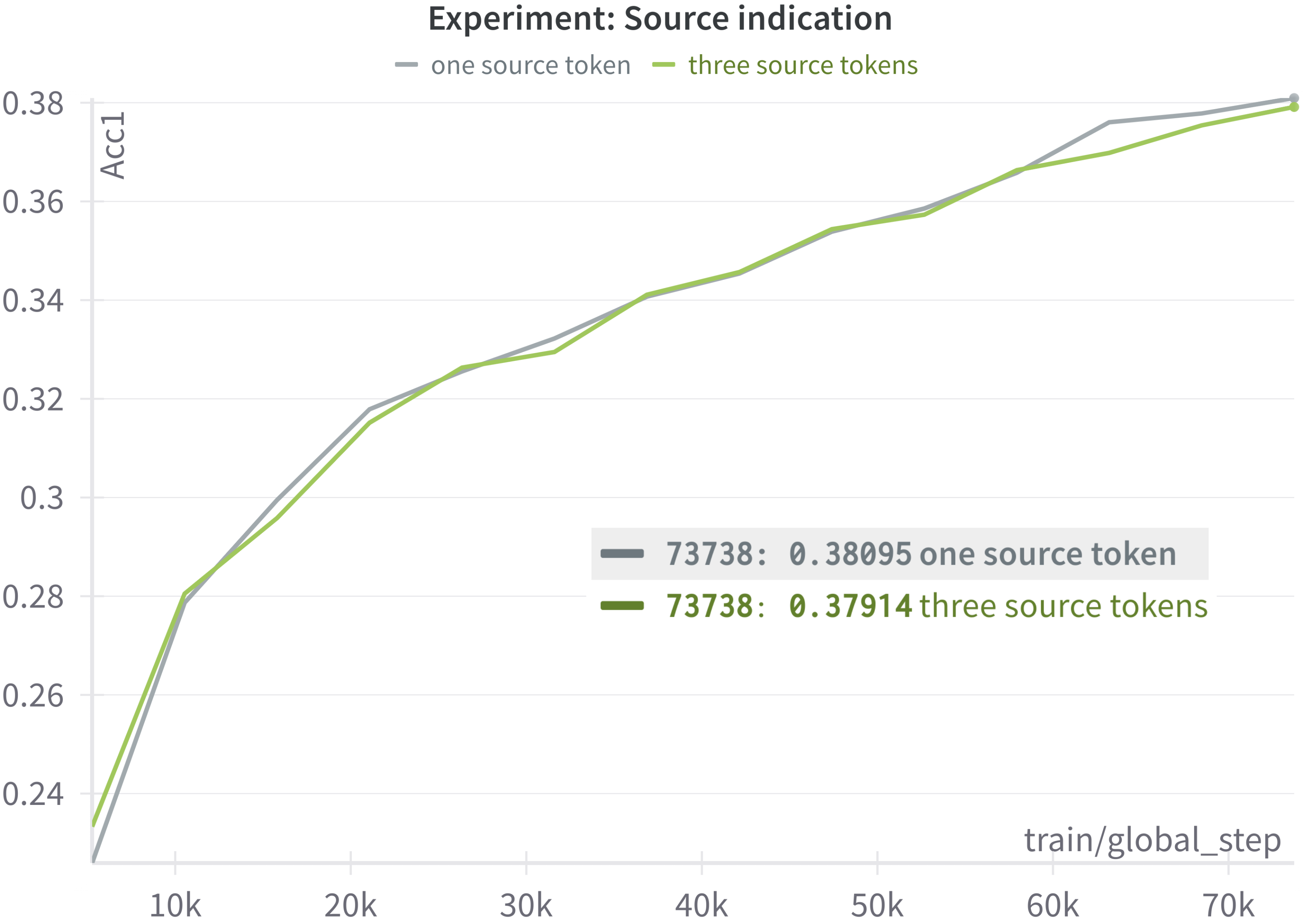}
    
    \caption{Tracked control evaluations of the \textbf{source indication} experiment. 
    Displays the dependency of Acc$_1$ on global step measured on the full NIST validation set. The table shows the final Acc$_1$ values.}
    \label{fig:exp_source_ind_acc}
\end{figure}

\clearpage\newpage
\subsection{Experiment 5: Training length and dataset size}
\label{app:experiment_train_len}

\begin{figure}[h]
    \centering
    \includegraphics[width=0.8\textwidth]{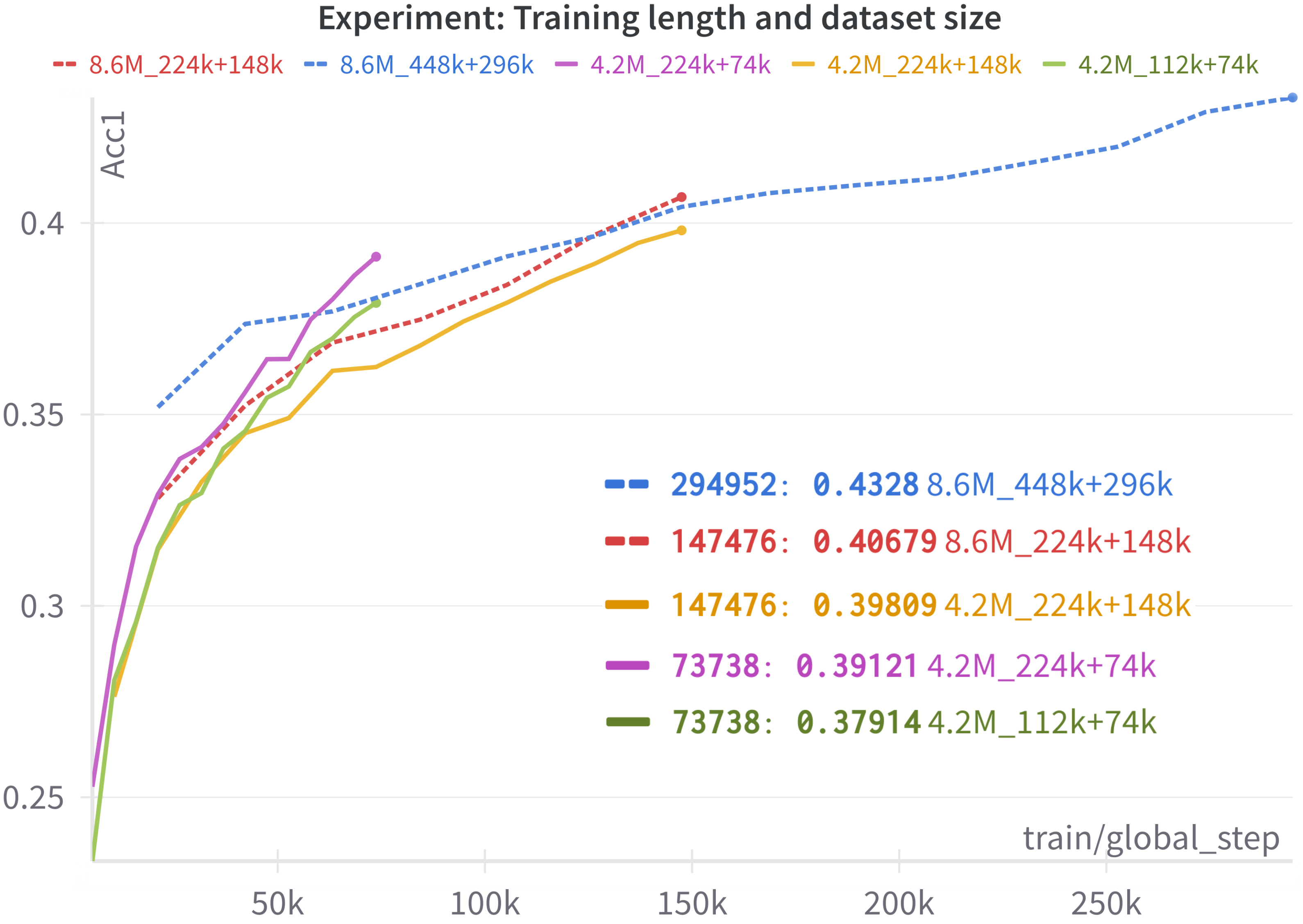}
    
    \caption{Tracked control evaluations of the \textbf{training length and dataset size} experiment. 
    Displays the dependency of Acc$_1$ on global step measured on the full NIST validation set. The table shows the final Acc$_1$ values.}
    \label{fig:exp_train_len_acc}
\end{figure}






\end{appendices}


\end{document}

%% file: tables/db_searches.tex



\begin{table}[h!]
    \centering
    \begin{tabular}{l|l|l|l|l}
    
        ~ &NIST & SWGDRUG & Cayman & MONA \\ \hline
        HSS$_{50}$ & 45.0\% & 55.2\% & 58.2\% & 22.2\% \\ \hline
        HSS$_{10}$ & 35.5\% & 47.1\% & 50.3\% & 16.0\% \\ \hline
        HSS$_{1}$ & 18.8\% & 21.0\% & 23.9\% & 5.9\% \\ \hline
        SSS$_{50}$ & 29.4\% & 29.3\% & 26.4\% & 16.1\% \\ \hline
        SSS$_{10}$ & 23.2\% & 23.2\% & 23.5\% & 11.7\% \\ \hline
        SSS$_{1}$ & 12.5\% & 9.5\% & 9.4\% & 5.2\% \\ \hline
    \end{tabular}
    \caption{Percentage of cases where database search methods (SSS and HSS) successfully retrieved the closest structure from the reference database among the top-1, top-10, and top-50 suggested candidates. Performance is evaluated across all test sets: NIST test split, SWGDRUG, Cayman, and MONA.}
    \label{tab:db_search_comparison}
\end{table}


%% file: tables/final_model_eval.tex


\begin{table}[h!]
    \centering
    \begin{tabular}{l|l|l|l|l|l|l|l|l}
        ~ & \multicolumn{2}{c|}{NIST} & \multicolumn{2}{c|}{SWGDRUG} & \multicolumn{2}{c|}{Cayman} & \multicolumn{2}{c}{MONA} \\
        ~ & Sim$_k$ & Acc$_k$ & Sim$_k$ & Acc$_k$ & Sim$_k$ & Acc$_k$ & Sim$_k$ & Acc$_k$ \\ \hline \hline
        BDC & 0.72 & 0.0\% & 0.68 & 0.0\% & 0.68 & 0.0\% & 0.70 & 0.0\% \\ \hline
        HSS$_{50}$ & 0.62 & 0.0\% & 0.61 & 0.0\% & 0.61 & 0.0\% & 0.47 & 0.0\% \\ \hline
        HSS$_{10}$ & 0.57 & 0.0\% & 0.58 & 0.0\% & 0.57 & 0.0\% & 0.42 & 0.0\% \\ \hline
        HSS$_1$ & 0.45 & 0.0\% & 0.46 & 0.0\% & 0.46 & 0.0\% & 0.27 & 0.0\% \\ \hline
        SSS$_{50}$ & 0.55 & 0.0\% & 0.50 & 0.0\% & 0.47 & 0.0\% & 0.44 & 0.0\% \\ \hline
        SSS$_{10}$ & 0.50 & 0.0\% & 0.46 & 0.0\% & 0.43 & 0.0\% & 0.38 & 0.0\% \\ \hline
        SSS$_1$ & 0.39 & 0.0\% & 0.34 & 0.0\% & 0.34 & 0.0\% & 0.26 & 0.0\% \\ \Xhline{3\arrayrulewidth}
        SpecTUS$_{50}$ & 0.84 & 69.8\% & 0.86 & 64.6\% & 0.78 & 45.2\% & 0.58 & 37.2\% \\ \hline
        SpecTUS$_{10}$ & 0.81 & 65.0\% & 0.82 & 58.5\% & 0.74 & 38.8\% & 0.54 & 34.0\% \\ \hline
        SpecTUS$_1$ & 0.67 & 43.3\% & 0.69 & 35.0\% & 0.60 & 20.7\% & 0.41 & 20.8\% \\ \hline
    \end{tabular}
    \caption{Comparison of Sim$_k$ and Acc$_k$ metrics between the final SpecTUS model and database search methods across all testing datasets (NIST test, SWGDRUG, Cayman, MONA).}
    \label{tab:final_model_comparison}
\end{table}

%% file: tables/win_alag.tex
\begin{table}[h!]
    \centering
    \begin{tabular}{l|l|l|l}

        ~ & BDC & HSS & SSS \\ \hline
        SpecTUS$_1$ & 46.8\% \textbf{/} 50.8\%& 76.4\% \textbf{/} 83.6\%& 79.6\% \textbf{/} 88.1\%\\ \hline
        SpecTUS$_{10}$ & 69.9\% \textbf{/} 72.2\%& 84.4\% \textbf{/} 87.3\%& 87.8\% \textbf{/} 91.2\%\\ \hline
        SpecTUS$_{50}$ & 75.8\% \textbf{/} 77.9\%& 85.6\% \textbf{/} 88.0\%& 88.9\% \textbf{/} 91.9\%\\ \hline
    \end{tabular}
    \caption{\textit{Win Rate} \textbf{/} \textit{At-least-as-good Rate} of SpecTUS over database search methods on the \textbf{NIST test} set. For all values, the compared performance involves the same number of candidates, e.g., $\text{ALAG}(\text{SpecTUS}_{10}, \text{HSS}_{10}$).}
\end{table}

\begin{table}[h!]
    \centering
    \begin{tabular}{l|l|l|l}
    
        ~ & BDC & HSS & SSS \\ \hline
        SpecTUS$_1$ & 44.8\% \textbf{/} 49.5\%& 72.3\% \textbf{/} 81.2\%& 87.7\% \textbf{/} 90.2\%\\ \hline
        SpecTUS$_{10}$ & 70.2\% \textbf{/} 73.7\%& 81.3\% \textbf{/} 84.9\%& 88.9\% \textbf{/} 92.3\%\\ \hline
        SpecTUS$_{50}$ & 77.5\% \textbf{/} 80.7\%& 85.3\% \textbf{/} 88.0\%& 90.9\% \textbf{/} 93.1\%\\ \hline
    \end{tabular}
    \caption{\textit{Win Rate} \textbf{/} \textit{At-least-as-good Rate} of SpecTUS over database search methods on the \textbf{SWGDRUG} dataset. For all values, the compared performance involves the same number of candidates, e.g., $\text{ALAG}(\text{SpecTUS}_{10}, \text{HSS}_{10}$).}
\end{table}

\begin{table}[h!]
    \centering
    \begin{tabular}{l|l|l|l}
    
        ~ & BDC & HSS & SSS \\ \hline
        SpecTUS$_1$ & 31.3\% \textbf{/} 35.0\% & 63.1\% \textbf{/} 67.4\% & 77.0\% \textbf{/} 81.7\% \\ \hline
        SpecTUS$_{10}$ & 56.3\% \textbf{/} 60.3\%& 71.0\% \textbf{/} 74.2\%& 82.7\% \textbf{/} 84.6\%\\ \hline
        SpecTUS$_{50}$ & 64.2\% \textbf{/} 68.7\%& 74.2\% \textbf{/} 77.2\%& 84.2\% \textbf{/} 86.4\%\\ \hline
    \end{tabular}
    \caption{\textit{Win Rate} \textbf{/} \textit{At-least-as-good Rate} of SpecTUS over database search methods on the \textbf{Cayman} library. For all values, the compared performance involves the same number of candidates, e.g., $\text{ALAG}(\text{SpecTUS}_{10}, \text{HSS}_{10}$).}
\end{table}

\begin{table}[h!]
    \centering
    \begin{tabular}{l|l|l|l}
    
        ~ & BDC & HSS & SSS \\ \hline
        SpecTUS$_1$ & 21.3\% \textbf{/} 23.4\%& 65.0\% \textbf{/} 70.6\%& 65.5\% \textbf{/} 72.4\%\\ \hline
        SpecTUS$_{10}$ & 35.6\% \textbf{/} 37.4\%& 63.7\% \textbf{/} 66.8\%& 66.8\% \textbf{/} 70.4\%\\ \hline
        SpecTUS$_{50}$ & 39.6\% \textbf{/} 41.6\%& 61.6\% \textbf{/} 64.9\%& 64.2\% \textbf{/} 68.0\%\\ \hline
    \end{tabular}
    \caption{\textit{Win Rate} \textbf{/} \textit{At-least-as-good Rate} of SpecTUS over database search methods on the \textbf{MONA} library. For all values, the compared performance involves the same number of candidates, e.g., $\text{ALAG}(\text{SpecTUS}_{10}, \text{HSS}_{10}$).}
    \label{tab:win_alag_nist}
\end{table}


%% file: tables/all_experiments.tex

\begin{table}[h!]
    \centering
    \begin{tabular}{l|l|l|l|l}
    
          & Acc$_1$ & Acc$_{10}$ & Sim$_1$ & Sim$_{10}$ \\ \hline
        exp5: 8.6M\_448k+296k (SpecTUS) & 43.3\% & 64.8\% & 0.67 & 0.81 \\ \hline
        exp5: 8.6M\_224k+148k & 40.7\% & 62.8\% & 0.65 & 0.80 \\ \hline
        exp5: 4.2M\_224k\_148k & 39.8\% & 61.8\% & 0.64 & 0.79 \\ \hline
        exp5: 4.2M\_224k\_74k & 39.1\% & 61.8\% & 0.64 & 0.79 \\ \hline
        exp4: one src token & 38.1\% & 61.4\% & 0.63 & 0.79 \\ \hline
        exp3: RASSP:NEIMS / exp5: 4.2M\_112k\_74k & 37.9\% & 60.7\% & 0.63 & 0.78 \\ \hline
        exp3: RASSP:NEIMS:NIST & 37.6\% & 60.6\% & 0.62 & 0.78 \\ \hline
        exp3: NEIMS-only & 36.8\% & 59.7\% & 0.62 & 0.77 \\ \hline
        exp3: RASSP-only & 34.8\% & 57.6\% & 0.61 & 0.76 \\ \hline
        exp1: log30bins / exp2: mf10M (char-level) / exp3: no pretraining & 28.1\% & 50.7\% & 0.56 & 0.72 \\ \hline
        exp1: log40bins & 27.6\% & 49.7\% & 0.55 & 0.72 \\ \hline
        exp1: lin4dec & 27.4\% & 49.9\% & 0.55 & 0.72 \\ \hline
        exp1: lin3dec & 27.3\% & 49.8\% & 0.55 & 0.72 \\ \hline
        exp1: log21bins & 27.2\% & 49.6\% & 0.55 & 0.72 \\ \hline
        exp1: log10bins & 24.7\% & 47.0\% & 0.54 & 0.70 \\ \hline
        exp2: mf10K & 24.5\% & 46.4\% & 0.54 & 0.70 \\ \hline
        exp2: mf100 & 24.5\% & 47.2\% & 0.55 & 0.71 \\ \hline
        exp2 mf10 & 24.0\% & 46.1\% & 0.54 & 0.70 \\ \hline
        exp1: lin2dec & 24.0\% & 46.1\% & 0.53 & 0.69 \\ \hline
        exp2: SELFIES & 22.3\% & 41.1\% & 0.49 & 0.65 \\ \hline
    \end{tabular}
    \caption{Summary of all experiments conducted during the development of the final SpecTUS model. The metrics were evaluated on the NIST validation set. The experiments are ordered by Acc$_1$ values, which served as the primary criterion for assessing the results.}
    \label{tab:experiments_comparison}
\end{table}